\documentclass[trsc,nonblindrev]{informs3arx} 

\OneAndAHalfSpacedXI 

\usepackage{natbib}
 \bibpunct[, ]{(}{)}{,}{a}{}{,}%
 \def\bibfont{\small}%
 \def\newblock{\ }%
\TheoremsNumberedThrough     

\EquationsNumberedThrough    

\usepackage{lineno,hyperref}
\usepackage[utf8]{inputenc}
\usepackage[]{mathtools}
\usepackage[]{multirow}
\usepackage[]{verbatim}
\usepackage[]{color}
\usepackage[]{rotfloat}
\usepackage[]{graphicx}
\usepackage[]{caption}
\usepackage[]{subcaption}
\usepackage[]{tikz}
\usepackage[]{booktabs}
\usepackage[]{paralist}

\newcommand{\eqflow}[1]{%
\sum_{j \in N} \left( x^r_{ij} - x^r_{ji} \right) + \sum_{\substack{h \in H \\ \text{if } i \in H}} \left(y^r_{ih} - y^r_{hi} \right)
      & #1 \begin{dcases*}
            1 & if $i = or(r)$ \\
            -1 & if $i = de(r)$ \\
            0 & otherwise
        \end{dcases*} & \forall r \in T, \forall i \in N}

\newcommand{\eqflowf}[1]{%
\sum_{j \in H} \left( x^r_{ij} - x^r_{ji} \right) + \sum_{\substack{h \in H \\ \text{if } i \in H}} \left(y^r_{ih} - y^r_{hi} \right)
      & #1 \begin{dcases*}
            1 & if $i = or(r)$ \\
            -1 & if $i = de(r)$ \\
            0 & otherwise
        \end{dcases*} & \forall r \in T, \forall i \in N}

\newcommand{\includemap}[1]{\includegraphics[width=\textwidth, clip, trim=660 0 690 0]{#1}}

\newcommand{\model}[2]%
    {\begin{subequations}%
        \label{mod:#1}%
        \begin{align}%
        #2
        \end{align}%
    \end{subequations}}
    
\newcommand{\named}[6]%
    { \begin{subequations}%
        \label{mod:#1}%
        \begin{align}%
        #3   && #4 \span \tag{#2} \\
        s.t. && #5 \\
             && #6 \span \nonumber
        \end{align}%
    \end{subequations}}
    
    {\begin{table}[htb]
    \centering
    \caption{#1}
    \label{tab:#2}
    \begin{tabular}{#3}
    \toprule}%
    {\bottomrule
    \end{tabular}
    \end{table}}

\newcommand{\Y}{\mathbb{Y}}
\newcommand{\B}{\mathbb{B}}
\newcommand{\R}{\mathbb{R}}

\DeclarePairedDelimiterX\Set[1]\{\}{%
    
    #1
}

\newcommand{\busplus}{{\sc BusPlus}}

\begin{document}
\RUNAUTHOR{Mah\'{e}o, Kilby, and Van Hentenryck}
\RUNTITLE{Benders Decomposition for a Hub and Shuttle Public Transit System}

\TITLE{Benders Decomposition for the Design of \\
a Hub and Shuttle Public Transit System}


\ARTICLEAUTHORS{%
\AUTHOR{Arthur Mah\'{e}o, Philip Kilby}
\AFF{The Australian National University, Acton ACT 2601, Australia \\
DATA61, Tower A, 7 London Circuit, Canberra City ACT 2601, Australia \\
 \EMAIL{\{Arthur.Maheo, Philip.Kilby\}@nicta.com.au}, \URL{}}
\AUTHOR{Pascal Van Hentenryck}
\AFF{The University of Michigan, Industrial and Operations Engineering \\
1205 Beal Avenue, Ann Arbor, MI 48109. \\ \EMAIL{pvanhent@umich.edu}, \URL{}}
}




\ABSTRACT{
  The \busplus{} project aims at improving the off-peak hours public
  transit service in Canberra, Australia. To address the difficulty of
  covering a large geographic area, \busplus{} proposes a hub and
  shuttle model consisting of a combination of a few high-frequency
  bus routes between key hubs and a large number of shuttles that
  bring passengers from their origin to the closest hub and take
  them from their last bus stop to their destination. 

  This paper focuses on the design of bus network and proposes an
  efficient solving method to this multimodal network design problem
  based on the Benders decomposition method. Starting from a MIP
  formulation of the problem, the paper presents a Benders
  decomposition approach using dedicated solution techniques for
  solving independent sub-problems, Pareto optimal cuts, cut bundling,
  and core point update. Computational results on real-world data from
  Canberra's public transit system justify the design choices and show
  that the approach outperforms the MIP formulation by two orders of
  magnitude. Moreover, the results show that the hub and shuttle model
  may decrease transit time by a factor of 2, while staying within the
  costs of the existing transit system.
}


\maketitle

\section{Introduction}

Canberra is a planned city designed by American architect Walter
Griffin in 1913. It features a large number of semi-autonomous towns
separated by greenbelts. As a result, Canberra covers a wide
geographic area, which makes public transportation particlarly
challenging. Bus routes are long and hence bus frequencies, and
patronage, are low, especially during off-peak periods. To address
these imitations, the \busplus{} project designed, optimized, and
simulated a Hub and Shuttle Public Transit System. The Hub and Shuttle
model consists of a combination of a few high-frequency bus routes
between key hubs and a large number of shuttles (or multi-hire taxis)
that bring passengers from their origin to the closest hub and take
them from their last bus stop to their destination. The main advantage
of \busplus{} is its ability to deliver the same service regardless of
the origin and destination of a passenger. \busplus{} uses bus stops
instead of a \emph{doorstep} shuttle service to reduce the delay of
waiting for the customer to exit her house, e.g., putting on shoes,
coat, ... In addition, bus stops provide an already established
network covering the city, within walking distance of homes even in
its remotest areas. From a traveller standpoint, this transit model is
highly convenient: Travellers book their travel online (e.g., on their
phones), are picked up at their traditional bus stop, and dropped at
their destination for the same ticket price as before. The anticipated
benefits come from the level of service: The hope is to reduce travel
time significantly for the same overall system cost.

Designing such a Hub and Shuttle Public Transit System (HSPTS) however
creates a series of interesting challenges, including
\begin{enumerate}
\item How to connect a set of potential hubs to minimize costs and
  maximize convenience?
\item How to allocate, during operations, requests to a trip,
  consisting of a shuttle leg, a number of bus legs, and a final
  shuttle leg?
\end{enumerate}

\noindent
This paper focuses on the first problem and primarily on off-peak
hours -- evenings, week-ends --, which are the most challenging from a
cost and service standpoint. Designing a HSPTS differs from the
traditional bus-network design problem which, given a set of known or
implied origin/destination demands, consists in building a network of
routes that visit all the bus stops and serve these demands. In
contrast, a solution to the HSPTS does not need to visit all the bus
stops nor even all the potential hubs. {\em The HSPTS goal is to
  decide which hubs to link via bus routes to take advantage of
  economies of scale, while relying on shuttles for the remaining
  "last mile" elements of the service.} As a result, in a HSPTS
system, the design of the bus network is intertwined with the routing
of passengers: The objective is to balance the cost of running buses
for a whole day, which has a high up-front cost, with the cost of
routing passengers with shuttles only, which has a low fixed cost but
incurs a costly per trip expense.

The paper models the HSPTS design problem as an Hub-Arc Location
Problem (HALP) \citep{Campbell2005,Campbell2005a} and uses Benders
decomposition \citep{Benders1962} to solve the resulting MIP
formulation. The Benders decomposition makes use of the Pareto-optimal
cuts proposed by \cite{magnanti1981accelerating}, core point updates,
and cut bundling using the problem structure. The experimental
results, based on real data collected on the Canberra public transit
system, show the benefits of the HSPTS proposed by the \busplus{}
project, as well as the effectiveness of Benders decomposition and its
various enhancements. Observe also that, although the main motivation
of the \busplus{} project concerns the off-peak setting, the
experimental results are based on actual data covering a whole day of
operations to validate the scalability of the model.

The rest of the paper is organized as follows. Section
\ref{sec:modeling} describes the HSPTS model considered in this paper,
its simplifying assumptions, and the data sets. Section
\ref{sec:literature} reviews prior work on related problems. Section
\ref{sec:mip} reviews the MIP formulation and a number of
problem-specific pre-processing techniques. Section
\ref{sec:introBenders} reviews Benders decomposition and some of its
important extensions.  Section \ref{sec:Benders} presents the Benders
decomposition approach. Finally, Section \ref{section:results}
presents the experimental results for the case study, including the
benefits of the Benders decomposition and the impact of the \busplus{}
project on the public transit system in Canberra.

In the following, we use shuttles and taxis interchangeably, since the
shuttles in our case study are multi-hire taxis, which are available
in large numbers in Canberra.

\section{Modeling the HSPTS Problem for the BusPlus Project}
\label{sec:modeling}

The HSPTS is a variant of the hub-arc location problem in which
customers have access to a multimodal transportation where buses run
on a fixed network and taxis can transport customers on any
origin-destination pair. Bus routes can be opened for a fixed cost
which represents the cost of operating high-frequency buses along the
arc. The goal is to minimize the cost of system, i.e., the fixed cost
of operating the bus lines and the variable cost for each taxi
trip, together with maximizing the convenience for the travellers. We
use the trip duration as a proxy for traveller convenience in the
model.

To make the design of the HSPTS manageable from a computational
standpoint, a number of simplifications are introduced. First, trips
are modeled as single commodities and the number of passengers is a
factor on the cost of the trip. Second, the routing and scheduling
aspects are ignored. Taxis are considered to be available at any
location within a short time and always travel directly from pickup to
setdown. This assumption is reasonable whenever the HSPTS system is
able to call on a sufficiently large taxi fleet, which is certainly
the case in Canberra. Buses are considered to be available for
connections with a nominal waiting time. The model does not account
for capacities, which is again a reasonable assumption in the off-peak
times. Obviously, the \busplus{} project has also developed online
optimization algorithms for taxi dispatching but this is outside
the scope of this paper.

\begin{figure}[t]
\centering
    \begin{subfigure}{0.45\columnwidth}
        \includemap{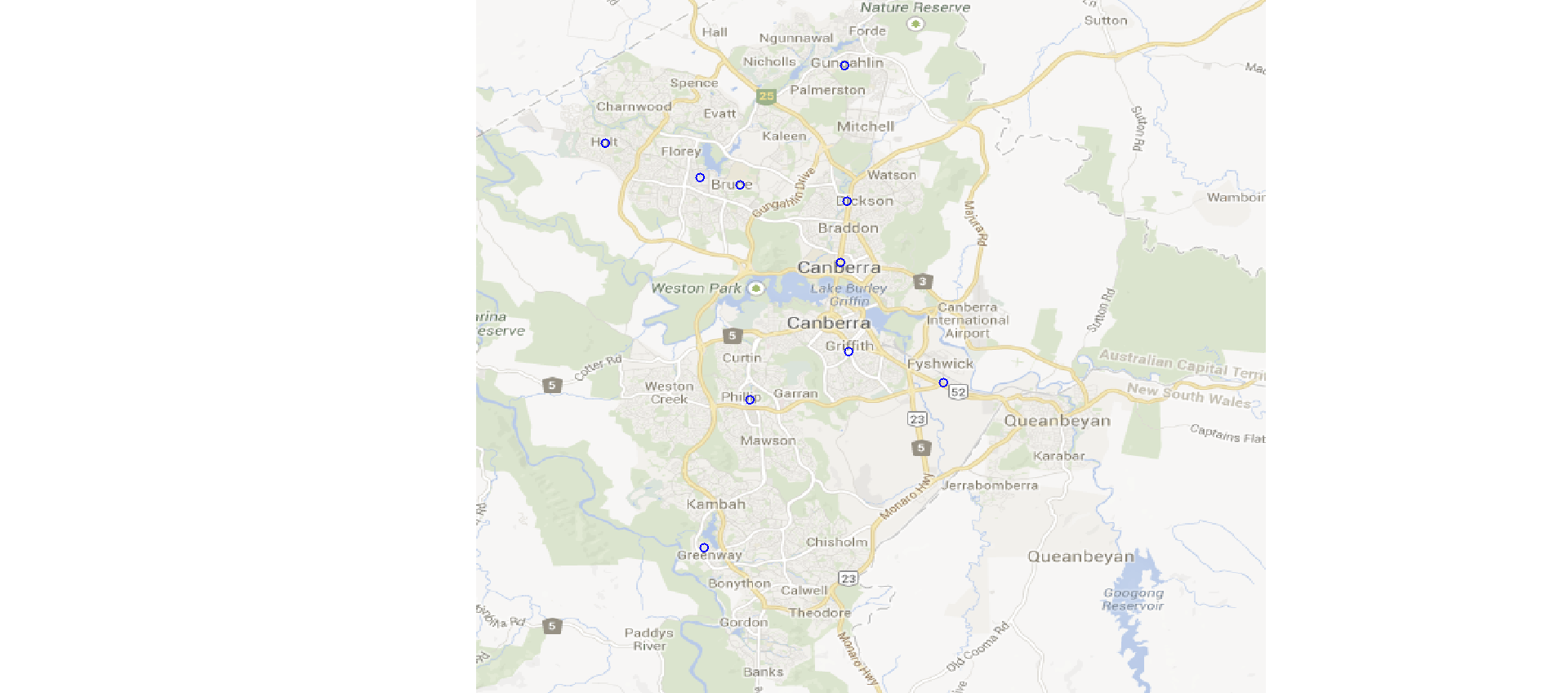}
        \subcaption{$H_{10}$}
    \end{subfigure}%
    \hfill
    \begin{subfigure}{0.45\columnwidth}
        \includemap{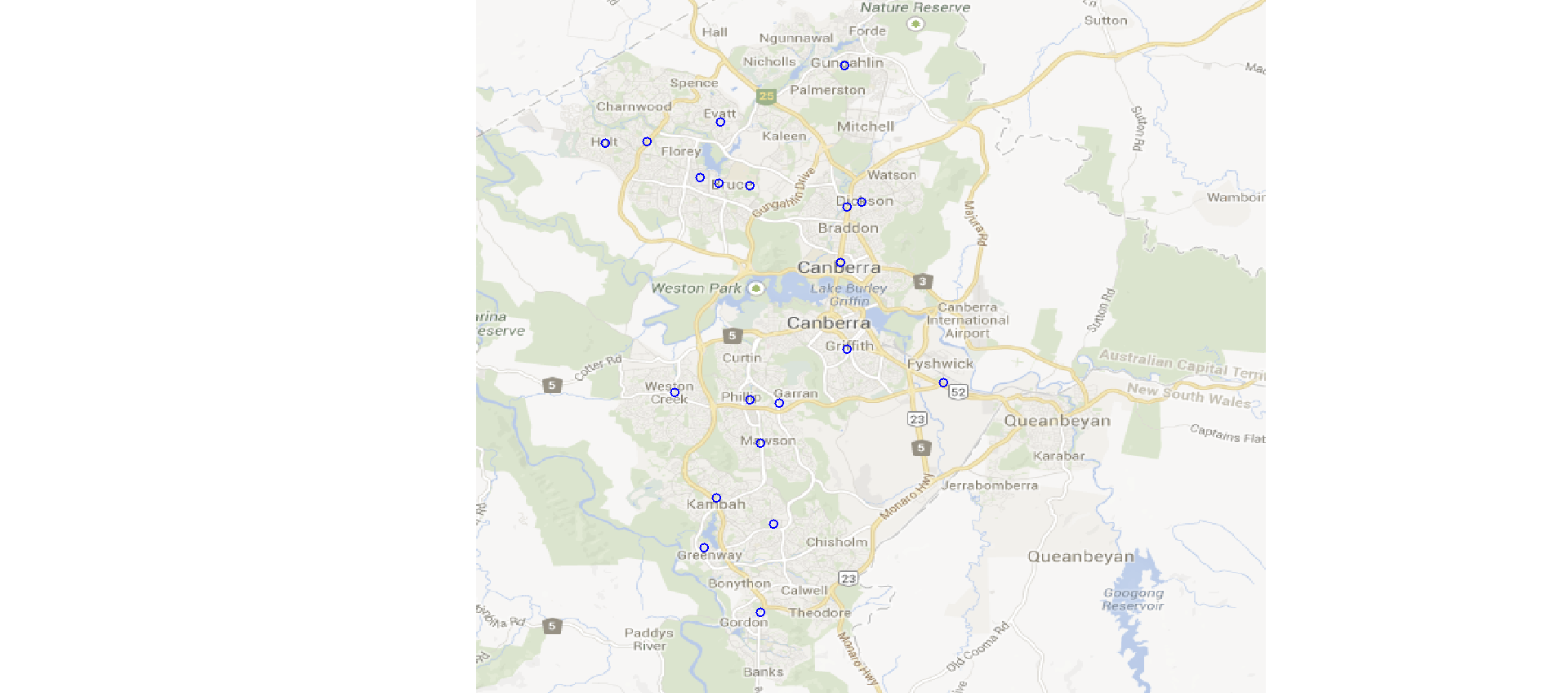}
        \subcaption{$H_{20}$}
    \end{subfigure}
    \caption{Two Potential Configurations for \busplus{} with 10 and 20 Hubs Respectively.}
    \label{fig:empty}
\end{figure}

Our dataset represents a week worth of trips in Canberra using the
current public transit network. Each trip has an origin and a
destination and a number of passengers. Time and distance matrices
between each pair of nodes gives the on-road distance and average
travel time between each pair of nodes: They both respect the triangle
inequality. Finally, Figure \ref{fig:empty} presents the two sets of
potential hubs to be used for building the bus network: $H_{10}$ with
ten hubs and $H_{20}$ with twenty hubs. The hubs are shown by blue
dots in the figure. The hubs were chosen using a \emph{p-median} model
\citep{hakimi1964optimum}.

\section{Review of Prior Work}
\label{sec:literature}

The problem of linking a set of hubs using arcs is called the Hub-Arc
Location Problem or \emph{hub-and-spoke} network design problem: It
was introduced by \cite{Campbell2005,Campbell2005a} and is defined as
the problem of locating a given number of hub arcs in such a way that
the total flow cost is minimized. It is mostly used in transhipment
contexts where economies of scale can be expected by grouping
flows. The Hub-Arc Location Problem (HALP) can be seen as a two-level
decision problem deciding which arcs to open first and then how to
route the flow at minimum cost. As such, its structure appears ideally
suited for Benders decomposition \citep{Benders1962}. In recent years,
a significant body of research has been dedicated to solving variants
of the Hub Location Problem and the HALP using Benders
Decomposition. These studies typically impose restrictions on the
network topology. They include, for instance, the Tree Hub Location
Problem \citep{de2013improved}, where hubs are connected in a
non-directed tree, the Hub Line Location Problem \citep{de2015exact},
where the decision is to locate $q$ lines connecting hubs, and star
topologies \citep{labbe2008solving,yaman2008star} where all hubs are
connected to a central hub and nodes are connected to hubs.
\cite{DeSa2015a} introduced the Hub Line Location Problem which
consists of locating $p$ hubs and connecting them by means of a path,
i.e., a single line. The arcs linking the hubs are a way to speed up
travel time but commodities can also be routed outside the
network. There may also be a cost to access the network. In order to
solve this complex problem, \cite{DeSa2015a} used a Benders
decomposition that include multicuts, a dedicated algorithm for the
sub-problem, and the use of a branch-and-cut framework.

\cite{Marin2009} solve a rapid urban transit network design problem,
where the goal is to locate train stations in an uncapacitated network
with line location and demand routing constraints. Moreover, customers
have access to a \emph{private} network in competition with the public
network of trains. They improve the convergence speed of the Benders
algorithm by using sub-problem disaggregation, a procedure to remove
inactive optimality cuts, and a dedicated sub-problem algorithm. The
goal in this work is to maximize the demand served by the transit
system, while our goal is to optimize a combination of cost and
convenience.

Finally, it is useful to mention that nodes can be assigned to hubs in
two main ways: single or multiple allocations. With multple
allocations, the commodities from a node can be sent to multiple
hubs. While the single allocation is the most commonly studied,
\cite{DeCamargo2008} apply Benders decomposition to the uncapacitated
multiple-aallocation HALP. The main issue is the non-linearity of
their model. As a result, they use a Generalised Benders Decomposition
with a MIP as the master problem and a non-linear convex
transportation problem as the subproblem. Note that \cite{DeSa2015a}
also allow for multiple allocations.

\section{A MIP Model}
\label{sec:mip}

This section presents a MIP model for the HSPTS problem. It starts by
presenting the main formulation and then proposes a number of
domain-specific pre-processing steps.

\subsection{The Basic Formulation}

\subsubsection{Inputs}

The inputs to the HSPTS are as follows:
\begin{enumerate}
\item a complete graph $G$ with a set $N$ of nodes;
\item a set $T$ of trips to serve, where each trip $r \in T$ is
  specified by a tuple $(o, d, p)$ with an origin $o^r \in N$, a
  destination $d^r \in N$, and a number of passengers $p^r \in
  \mathbb{N}$;
\item a subset $H \subseteq N$ of nodes that can be used as bus hubs.
\end{enumerate}  
\noindent
In the following, we use $or(r)$ and $de(r)$ to denote the origin and
destination of a trip $r$.  The distance and the travel time from node
$i$ to $j$ ($i$, $j \in N)$ are given by $d_{ij}$ and $t_{ij}$
respectively. Since the HSPTS problem aims at optimizing cost and
convenience jointly, we use $\alpha$ as a \emph{conversion factor} to
translate travel times into financial costs. Travel distance is used
as a cost measure, while travel time is a proxy for convenience. The
value $\alpha$ shifts the solution towards cost or convenience by
modifying travel times by a factor $\alpha$ and travel costs by $(1 -
\alpha)$: The lower $\alpha$ is, the more cost-driven the solution
is. The objective is formulated in terms of the following constants:
\begin{itemize}
\item[$c$:] the cost of using a taxi per kilometer;
\item[$b$:] the cost of using a bus per kilometer;
\item[$n$:] the number of buses per day which depends on the headway;
\item[$S$:] the average waiting for a bus.
\end{itemize}

\noindent
The different characteristics of the two transportation modes are
captured in their associated cost functions: The cost ($\tau$) of a
taxi is a combination of a cost per kilometer and a cost per minute,
while the cost ($\gamma$) of bus is only a function of time. However,
buses run for the whole day, which is modelled by a large initial
set-up cost ($\beta$). These costs are defined as
follows: \begin{itemize}
\item $\tau_{ij} = (1 - \alpha) c \cdot d_{ij} + \alpha \cdot t_{ij}$ \hfill
              (cost and convenience of using a taxi from $i$ to $j$);
        \item $\gamma_{hl} = \alpha (t_{hl} + S)$ \hfill
              (convenience of using a bus from $h$ to $l$);
        \item $\beta_{hl} = (1 - \alpha) b \cdot n \cdot d_{hl}$ \hfill
              (cost of opening a bus leg from $h$ to $l$).
    \end{itemize}
    
\noindent
Since all the passengers in a trip $r$ are travelling at the same
time, the taxi and bus costs of $r$ are obtained by multiplying the
arc costs by the number of passengers, i.e.,
\begin{itemize}
\item $\tau^r_{ij} = p^r \cdot \tau_{ij};$
\item $\gamma^r_{hl} = p^r \cdot \gamma_{ij}.$
\end{itemize}

\subsubsection{Decision Variables} 
The decision variables for the HSPTS problem, which are all binary,
are as follows: Binary variable $x^r_{ij}$ denotes whether trip $r$
uses a taxi to travel arc $(i, j)$, variable $y^r_{hl}$ denotes
whether trip $r$ uses a bus on arc $(h, l)$ $(h, l \in H)$, and
variable $z_{hl}$ indicates whether arc $(h, l)$ is opened for buses
to use. By convention, this paper uses $(i, j)$ for edges travelled by
taxis and $(h, l)$ for edges used by buses.

%
%

\subsubsection{Network Topology}
        
The MIP model only enforces a weak form of connectivity that requires
that the sum of all incoming bus legs must be equal to the sum of all
outgoing bus legs at every hub. In other words, each leg entering a
hub has a corresponding leg which has to leave the hub. While this
formulation technically leaves open the possibility of producing a
network of disconnected components, the demand patterns observed in
practice lead to networks of good quality. With low demand, the
topology is often a simple circuit around the center. Higher demand
patterns often produces one or more connected flower topologies with
sub-circuits extending from the center to the suburbs. Figure
\ref{fig:networks} in Section \ref{section:results} illustrates these
topologies.
    
\subsubsection{The MIP Model}

We are now in a position to present the MIP model for the HSPTS of
\busplus{}. This model has similarities with the HAL4 from
\cite{Campbell2005} but it relaxes a number of critical assumptions
for our case study, i.e.,
\begin{enumerate}
\item An optimal solution may contain a path from the origin to the destination that does not contain a hub.
\item There is no constraint on the number of arcs to open;
\item The model allows for bridge arcs between two hubs.
\end{enumerate}

\begin{figure}[t]
\named{mip}%
    {MIP}%
    {\min}%
    {\sum_{r \in T} \sum_{i,j \in N} \tau_{ij}^r x_{ij}^r + \sum_{r \in T}\sum_{h, l \in H} \gamma_{hl}^r y^r_{hl} + \sum_{h,l \in H} \beta_{hl} z_{hl} \span}%
    {\sum_{l \in H} z_{hl} & = \sum_{l\in H} z_{lh} & \forall h \in H \label{eq:hub} \\
     & & y^r_{hl} & \leq z_{hl} & \forall r\in T, \forall h,l \in H \label{eq:hard} \\
     & & \eqflow{=} \label{eq:flow}}%
    {x_{ij}^r, y_{hl}^r, z_{hl} \in \B}
\caption{The MIP Model for the \busplus{} HSPTS.}
\label{fig:mip}
\end{figure}

\noindent
The MIP model is presented in Figure \ref{fig:mip}. The objective
function is the sum of the travelling cost of the trips using their
selected arcs and the cost of opening the bus legs. Constraints
(\ref{eq:hub}) enforce the weak connectivity constraints on the bus
legs, Constraints (\ref{eq:hard}) ensure that travellers only use
opened bus legs, and Constraints (\ref{eq:flow}) enforce flow
conservation, i.e., travellers start at their origin and reach their
destination without skipping network edges.  Note that, once the $z$
variables are fixed, the problem becomes totally unimodular and
the integrality constraints on the $x$ and $y$ variables can be
relaxed.


\subsection{Problem-Specific Preprocessing}
\label{sec:opti}

We now describe two filtering techniques that decrease the number of
decision variables significantly: Trip filtering and link filtering.

\subsubsection{Trip Filtering}

It may be the case that, in all possible configurations of the HSPTS
problem, the optimal routing of a trip $r$ is a direct taxi
ride. These trips can be filtered from the HSPTS problem, since they
do not impact its optimal solutions. Such trips can be identified by a
simple filtering algorithm that
\begin{enumerate}
\item considers that all bus legs are open;
\item computes a least-cost path from the source to the destination of each trip;
\item removes the trip if the least-cost path is a direct taxi ride.
\end{enumerate}
    
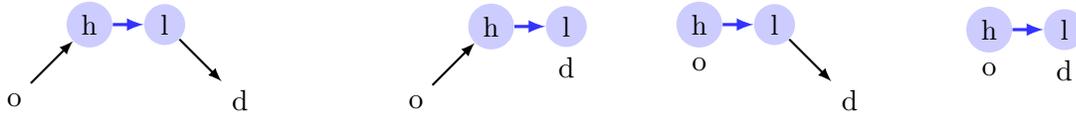
\begin{figure}[t]
\centering
\begin{subfigure}{0.3\textwidth}
\begin{tikzpicture}[node/.style={fill=none,draw=none}]
    \node (o) at (0,0) [] {o};
    \node (h) at (1,1) [circle,fill=blue!20,scale=0.75] {\large h};
    \node (l) at (2,1) [circle,fill=blue!20,scale=0.75] {\large l};
    \node (d) at (3,0) [] {d};
    \draw[-latex,thick] (o) -- (h);
    \draw[-latex,very thick,blue!80] (h) -- (l);
    \draw[-latex,thick] (l) -- (d);
\end{tikzpicture}
\end{subfigure} %
~ %
\begin{subfigure}{0.2\textwidth}
\begin{tikzpicture}[node/.style={fill=none,draw=none}]
    \node (o) at (0,0) [] {o};
    \node (h) at (1,1) [circle,fill=blue!20,scale=0.75] {\large h};
    \node (l) at (2,1) [circle,fill=blue!20,label=below:d,scale=0.75] {\large l};
    \draw[-latex,thick] (o) -- (h);
    \draw[-latex,very thick,blue!80] (h) -- (l);
\end{tikzpicture}
\end{subfigure} %
~ %
\begin{subfigure}{0.21\textwidth}
\begin{tikzpicture}[node/.style={fill=none,draw=none}]
    \node (h) at (0,1) [circle,fill=blue!20,label=below:o,scale=0.75] {\large h};
    \node (l) at (1,1) [circle,fill=blue!20,scale=0.75] {\large l};
    \node (d) at (2,0) [] {d};
    \draw[-latex,very thick,blue!80] (h) -- (l);
    \draw[-latex,thick] (l) -- (d);
\end{tikzpicture}
\end{subfigure} %
~ %
\begin{subfigure}{0.2\textwidth}
\begin{tikzpicture}[node/.style={fill=none,draw=none}]
    \node (dummy) at (0, 0) {};
    \node (h) at (0,1) [circle,fill=blue!20,label=below:o,scale=0.75] {\large h};
    \node (l) at (1,1) [circle,fill=blue!20,label=below:d,scale=0.75] {\large l};
    \draw[-latex,very thick,blue!80] (h) -- (l);
\end{tikzpicture}
\end{subfigure} %
\caption{Possible Trip Patterns (Not Showing Direct Taxi Trips).}
\label{fig:trips}
\end{figure}

\noindent
It is interesting to observe that, in the filtering procedure, a
least-cost trip can be a direct taxi ride or one of the four patterns
depicted in Figure \ref{fig:trips}, since it is assumed that all bus legs
are open. In particular, a least-cost trip is either (1) a single taxi
ride or (2) a journey with exactly one bus leg and possibly a taxi
ride from the origin to the hub and/or from the hub to the
destination.

\begin{table}[t]
\centering
 \begin{tabular}{l  c *{2}{| c c}}
 \hline
    & & \multicolumn{2}{c|}{$H_{10}$} & \multicolumn{2}{c}{$H_{20}$} \\
    & Total & T-Filtered & Reduction (\%) & T-Filtered & Reduction (\%) \\
    \midrule
    Monday     & 21282 & 14324 & 32.69 & 17294 & 18.74 \\
    Tuesday    & 21029 & 14184 & 32.55 & 17084 & 18.76 \\
    Wednesday  & 21418 & 14451 & 32.53 & 17401 & 18.76 \\
    Thursday   & 21487 & 14486 & 32.58 & 17499 & 18.56 \\
    Friday     & 19809 & 13398 & 32.36 & 16013 & 19.16 \\
    \hline
 \end{tabular}
 \caption[Number of Trips]{Effectiveness of Trip Filtering. Column Total gives the initial number of trips. Columns T-Filtered report the number of trips after trip filtering. Columns Reduction give the trip reduction in percentage.}
 \label{tab:warmed}
\end{table}

Table \ref{tab:warmed} reports experimental results on the effectiveness of
trip filtering. Column \emph{Total} reports the initial number of
trips, while the \emph{T-Filtered} columns give the trips remaining
after filtering for both hub configurations. Column \emph{Reduction}
report the trip reduction in percentage for both configurations. For
the 10 hub configuration, the filtering reduces the number of trips by
more than 30\%.  The reduction is more than 18\% in the configuration
with 20 hubs. This lower number was of course expected for 20 hubs,
since additional hubs make more complex hub patterns more attractive
from a cost standpoint.

\subsubsection{Link Filtering}
 
The current formulation considers every possible taxi link, thus using
a complete graph. Because of the triangle inequality, there is no need
to connect all the nodes using taxis: The only taxi links to consider
for a trip $r=(o,d,p)$ are

\begin{enumerate}
\item from the origin to a hub;
\item from a hub to the destination;
\item and from the origin to the destination.
\end{enumerate}
As a result, the formulation needs only the following taxi variables:
\begin{align*}
x^r_{oh} \geq 0 & \quad \forall r \in T, \forall h \in H; \\
x^r_{hd} \geq 0 & \quad \forall r \in T, \forall h \in H; \\
x^r_{od} \geq 0 & \quad \forall r \in T.
\end{align*}
    
\noindent
Observe that $\tau_{od}$, the cost of a one-person direct taxi trip,
is an upper bound to the cost of a trip. This upper bound can be used
to filter links from the origin $o$ to a hub $h$ and from a hub $l$ to
the destination $d$ by generalizing the trip filtering presented
earlier. Indeed, if the least-cost path going through $(o,h)$ is not
cheaper than $\tau^r_{od}$, variable $x^r_{oh}$ can be
removed. Consider, for instance, the case where the trips have three
components:
\begin{inparaenum}[1)]
\item a taxi trip from the origin to a hub;
\item a bus trip between two hubs; and
\item a taxi trip from the last hub to the destination.
\end{inparaenum}
If the condition
\begin{equation*}
\forall l: \tau_{oh} + \gamma_{hl} + \tau_{hl} > \tau_{od} 
\end{equation*}
holds, then the taxi trip $(o,h)$ will never be used and the variable
$x^r_{oh}$ can be removed. The symmetric condition
\begin{equation*}
\forall h: \tau_{oh} + \gamma_{hl} + \tau_{hl} > \tau_{od} 
\end{equation*}
allows to remove variable $x^r_{ld}$. Similar reasonings can be applied to
the other patterns depicted in Figure \ref{fig:trips}

 \begin{table}[t]
 \centering
    \begin{tabular}{l *{2}{| c c c}}
    \hline
         & \multicolumn{3}{c |}{$H_{10}$} & \multicolumn{3}{c}{$H_{20}$} \\
         & T-Filtered & L-Filtered & Reduction (\%)  & T-Filtered & L-Filtered & Reduction (\%) \\
         \hline
         Monday    & 292958 & 131580 & 55.09\% & 698910 & 294726 & 57.83\% \\
         Tuesday   & 290100 & 129998 & 55.19\% & 690278 & 291736 & 57.74\% \\
         Wednesday & 296000 & 133533 & 54.89\% & 706873 & 299117 & 57.68\% \\
         Thursday  & 295555 & 132500 & 55.17\% & 703205 & 296852 & 57.79\% \\
         Friday    & 273226 & 124953 & 54.27\% & 646095 & 279811 & 56.69\% \\
         \hline
     \end{tabular}
\caption[Number of Taxi Trips]{Effectiveness of Link Filtering: The T-Filtered and L-Filtered columns give the number of taxi variables left after applying the T-filtering and L-filtering procedures. The reduction columns give the percentage of taxi variables pruned by L-filtering from the T-filtered instances.}
\label{tab:groom}
 \end{table}

 We call this procedure link filtering (L-filtering for short). Once
the T-filtering and L-filtering procedures have been applied, the
remaining taxi arcs are called {\em useful} and $D^r$ denotes the set
of useful taxi arcs for a trip $r$. Table \ref{tab:groom} presents
experimental results on link filtering. They indicate that the link
filtering procedure is particularly effective, removing more than 50\%
of the taxi arcs available after trip filtering. As the table shows,
this holds for the two configurations with 10 and 20 hubs
respectively.

\section{A Brief Review of  Benders Decomposition}
\label{sec:introBenders}

This section provides a brief overview of Benders decomposition for
solving MIPs \citep{Benders1962}. The overview uses some simplyfying
assumptions that hold in the HSTPS problem. Assume, for simplicity,
that the MIP model is specified as follows:
\model{original}{
\min && c^T x + f^T y \span \\
s.t. && Ax + By & \geq b & ~\\
     && x \geq 0, y \in \Y \span
}

\noindent
where the $x$ variables are real-valued and the $y$ variables are
discrete and belong to the feasible set $\Y$.  Benders decomposition
splits the MIP model into two parts: a master problem that contains
the $y$ variables and a sub-problem that contains the $x$ variables
and a candidate solution for the $y$ variables. The subproblem can be 
specified as 
\model{subproblem}{
q(y) = \min && c^T x \span \\
s.t. && Ax & \geq b - B y & ~ \label{eq:prim_ax} \\
     && x \geq 0 \span
}

\noindent
and the master
problem becomes \model{primal}{
  \min && f^T y + q(y) & ~ & ~ \label{eq:master_obj} \\
  && y \in \mathbb{Y} \label{eq:master_domain} }

\noindent
For simplicity, we assume that the subproblem \eqref{mod:subproblem}
is always feasible and bounded, which will be the case for the HSTPS
problem. Let $\alpha$ be the dual variables associated with
\eqref{eq:prim_ax}. The dual of \eqref{mod:subproblem} is given by
\model{dual}{
  \max && \alpha^T (b - By) \span \\
  s.t. && A^T \alpha & \leq c & ~\\
  && \alpha \geq 0 \span }

\noindent
Let $\pi_j$ $(j \in [1, J])$ be the extreme points of the dual
feasibility region, which does not depend on the candidate values for
the $y$ variables. The subproblem \eqref{mod:subproblem} can be
reformulated as
\model{q}{
\min && q \span \\
s.t. && \pi_j^T (b - By) & \leq q & \forall j \in [1, J] \\
     && q \in \R \span
}
and the master problem as 
\model{master}{
\min && f^T y + q \span \\
s.t. && \pi_j^T (b - By) & \leq q & \forall j \in [1, J] \label{eq:optimality} \\
     && q \in \R, y \in \Y \span 
}
where the constraints \eqref{eq:optimality} are called Benders (optimality) \emph{cuts}.

Since there are potentially an exponential number of extreme points,
Benders decomposition starts by solving a restricted master problem
with no cuts and introduces the Benders cuts lazily. At each
iteration, Benders decomposition uses the restricted master to
produces candidate solution $(\bar{y},\bar{q})$ and then solves the
subproblem $q(\bar{y})$. If $q(\bar{y}) = \bar{q}$, the candidate
solution is optimal. Otherwise, a Benders cut of the form
\eqref{eq:optimality} is added to the restricted master problem and
the process is repeated.

\subsection{Separable Benders Subproblem and Cut Bundling}

Consider now the case where the subproblem \eqref{mod:subproblem} is separable, i.e., it can be rewritten as
\model{sum_sub}{
\min && c_1^T x_1 + c_2^T x_2 + \dots + c_n^T x_n \span \\
s.t. && A_1 x_1 & \geq b_1 - B_1 y_1 \\
     && A_2 x_2 & \geq b_2 - B_2 y_2 \\
     &&& \ddots \nonumber \\
     && A_n x_n & \geq b_n - B_n y_n \\
     && x_i \geq 0 \span & \forall i \in [1, n] 
      \nonumber
}
Solving the subproblem now consists of optimizing independent components, i.e.,
    $$
    \eqref{mod:sum_sub}\Leftrightarrow %
        \begin{aligned}
        \min && c_1^T x_1 \\
        s.t. && A_1 x_1 & \geq b_1 - B_1 y_1 \\
             && x_1 \geq 0
        \end{aligned} %
        + %
        \begin{aligned}
        \min && c_2^T x_2 \\
        s.t. && A_2 x_2 & \geq b_2 - B_2 y_2 \\
             && x_2 \geq 0
        \end{aligned} %
        + \dots
    $$
\noindent
A separable subproblem gives some new possibilities for cut
generation. Indeed, \cite{geoffrion1974multicommodity} observe that,
in this case, there exist better strategies than aggregating the
results of the sub-optimizations to produce a single cut (see also the
concept of {\em multi-cut} proposed by \cite{birge1988multicut}).  We
will investigate various \emph{cut bundling} strategies for the HSTPS
problem in the next section. 

\subsection{Pareto Optimal Cuts}

When the subproblem is a network flow optimization, which exhibits
high degeneracy \citep{ahuja1988network}, Benders decomposition may
experience slow convergence and it becomes important to generate
stronger, \emph{Pareto optimal}, cuts as proposed by
\cite{magnanti1981accelerating}. Given two solutions $\pi_a$ and
$\pi_b$ to \eqref{mod:subproblem}, $\pi_a$ dominates $\pi_b$ if
\begin{align*}
\pi^T_a (b - By) & \geq \pi^T_b (b - By),
\end{align*}
and at least one of these inequalities is strict. A solution (and its
associated cut) that dominates all other solutions is said to be
Pareto optimal. \cite{magnanti1981accelerating} showed how to generate a
Pareto optimal cut by solving an optimization problem, using the
notion of a \emph{core point} of a set $s$, i.e., a point in the
relative interior of the convex hull of $S$ denoted by $ri(S)$.  A
Pareto optimal cut can be obtained by solving the following
optimization problem:
\model{pareto}{
\max && \alpha^T (b - By^0) \span \\
s.t. && A^T \alpha & \leq c & ~\\
     && \alpha^T (b - B\bar{y}) & = q(\bar{\alpha}) \label{eq:pareto} \\
    && \alpha \geq 0 \span
}
\noindent
where $\bar{y}$ is the candidate solution from the restricted master
problem, $\bar{\alpha}$ is an optimal solution to \eqref{mod:dual},
and $y^0 \in ri(\Y)$ is a core point of $\Y$.

    
\cite{papadakos2009integrated} showed that the choice of a core point
can dramatically improve the convergence rate of Benders
decomposition. Moreover, the author proposed a method for computing a
sequence of core points by using a linear combination of the current
core point and the solution obtained at each iteration of Benders
decomposition. Experimental results have shown that using weights
of $1/2$ gives excellent results. 

\section{Benders Decomposition for the HSTPS Problem}
\label{sec:Benders}

This section presents how to apply Benders decomposition to the HSTPS
problem. The key idea is to assign the $z$ variables in the MIP model
to the restricted master problem, leaving the $x$ and $y$ variables
for the subproblem.  As mentioned earlier, when the $z$ variables are
fixed to $\bar{z}$, the subproblem is a min-cost flow, which can be
solved efficiently. The subproblem can be specified as
\model{bus_subpb}{  \min && \sum_{r \in T} \left( \sum_{i, j \in N}
    \tau_{ij}^r x_{ij}^r +%
    \sum_{h, l \in H} \gamma_{hl}^r y_{hl}^r \right)  \\
  s.t. && \eqflowf{=} \label{subproblem:flow} \\
  && y^r_{hl}  \leq \bar{z}_{hl} && \forall r \in T, h,l \in H \label{subproblem:link} \\
  && 0 \leq x_{ij}^r, y_{hl}^r \leq 1 \nonumber }
    
\noindent
The dual of the problem can be specified in terms of the dual
variables $u_i^r$ associated with constraints \eqref{subproblem:flow}
variable and the dual variables $v_{hl}^r$  associated with
constraints \eqref{subproblem:link}.

\model{bus_dual}{
    {\max} &&
    \sum_{r \in T} (u_o^r - u_d^r) - \sum_{h, l \in H} z_{hl} \left( \sum_{r \in T} v_{hl}^r \right) \\
    s.t. &&  u_i^r - u_j^r  \leq \tau^r_{ij} && \forall r \in T, i,j \in D^r \\
     && u^r_h - u^r_r - v^r_{hl}  \leq \gamma_{hl}^r && \forall r \in T, h,l \in H \\
     && u_i^r \geq 0, v_{hl}^r \geq 0 \nonumber }

\noindent
Note that Model \eqref{mod:bus_subpb} is separable: Each trip $r$ gives rise
to an independent subproblem. 
        
\subsection{Cut Aggregation}

We now discuss how to generate the cuts for the restricted master
problem. Since the subproblem is separable, their individual cuts can
be aggregated in many ways and we explored a variety of cut bundling
strategies that exploit the structure of the HSTPS problem.  For
instance, a bundling strategy may aggregate the cuts for all the trips
with the same origin. It is also important to mention that each
bundled cut must create its own variable and that each original
variable should appear in exactly one cut. The following cut bundling
strategies were investigated for the HSTPS problem (in the cut
definitions, $\bar{u}_i^r$ and $\bar{v}_{hl}^r$ denote the values of
the dual variables in the subproblem).

\begin{description}
\item[One] The traditional Benders cut from \eqref{mod:bus_subpb}.
        \begin{align}
        \label{eq:cut_1}
        \sum_{r \in T} (\bar{u}_{o}^{r} - \bar{u}_{d}^{r}) - %
            \sum_{h, l \in H} z_{hl} \left( \sum_{r \in T} \bar{v}_{hl}^{r} \right) &%
            \leq q & ~
        \end{align}
\item[Multi] This strategy does not aggregate the cuts.
        \model{all}{
        \bar{u}_{o}^{r} - \bar{u}_{d}^{r} - %
            \sum_{h, l \in H} z_{hl} \left( \sum_{r \in T} \bar{v}_{hl}^{r} \right) &%
            \leq q^r & \forall r \in T \\
        \sum_{r \in T} q_r \leq q
        }

    \item[Hubs] This strategy aggregates the subproblems by the closest hub to a trip origin or destination. Let $T_h$ be the set of trips having $h$ as their closest hub.
        \model{hub}{
        \sum_{r \in T_h} (\bar{u}_{o}^{r} - \bar{u}_{j}^{r}) - %
            \sum_{l \in H} z_{hl} \left( \sum_{r \in T_h} \bar{v}_{hl}^{r} \right) &%
            \leq q_h & \forall h \in H \\
        \sum_{h \in T_h} q_h \leq q
    }
    \item[Origin] This strategy aggregates the subproblems by trip origins. Let $T_o$ be trips having origin $o$ and $O$ be the set of all possible origins.
        \model{origin}{
        \sum_{r \in T_{o}} (\bar{u}_{o}^{r} - \bar{u}_{j}^{r}) - %
            \sum_{h, l \in H} z_{hl} \left( \sum_{r \in T_{o}} \bar{v}_{hl}^{r} \right) &%
            \leq q_o & \forall o \in O \\
        \sum_{o \in O} q_o \leq q
        }
      \item[Legs] This strategy aims at grouping the trips by the first
        bus leg they use. Since this bus leg is not known a priori,
        the leg strategy aggregates the subproblems by the bus leg
        used in trip filtering. Recall that all trips are associated
        with a single bus leg in trip filtering since all the bus legs
        are open. Let $T_{hl}$ be the set of trips using bus leg
        $(h,l)$ in the trip filtering.
        \model{leg}{ 
          \label{eq:cut_5}
          \sum_{r \in
            T_{hl}} (\bar{u}_{o}^{r} - \bar{u}_{j}^{r}) - %
          z_{hl} \sum_{r \in T_{hl}} \bar{v}_{hl}^{r} &%
          \leq q_{hl} & \forall h,l \in H \\
          \sum_{h,l \in H} q_{hl} \leq q 
        }
    \end{description}
    
    \subsection{Restricted Master Problem}

    We are now in a positon to present the restricted master problem,
    whose objective minimizes the cost of opening the bus legs subject
    to the circular constraint \eqref{eq:hub} and one of the cut sets
    defined above for each iteration:
    \model{bus_master}{
    \min\quad & \sum_{h,l \in H} \beta_{hl} z_{hl} + q  \nonumber \\
    s.t.\quad & \eqref{eq:hub}, \text{ one of } \eqref{eq:cut_1}-\eqref{mod:leg} \nonumber \\
              & z \in \B, q \in \R \nonumber
    }

\subsection{Pareto-Optimal Cuts}
        
Our Benders implementation for the HSTPS problem uses Pareto-optimal
cuts, which requires to solve two linear programs: The original
sub-problem \eqref{mod:bus_subpb} and then the Pareto sub-problem.

Observe first that each of the independent subproblems is a min-cost
flow with edges of infinite capacity and is equivalent to a shortest
path problem. Hence, the original subproblem for a given trip can be
solved by computing, for each trip, a shortest path between the origin
and the destination using the edges defined by the union of $D^r$, the
useful taxi arcs, and $\bar{Z}$, the set of opened bus legs in the
current iteration.

To define the Pareto subproblem, it is necessary to find a core point
that satisfies the circular constraint \eqref{eq:hub}. It can be chosen easily
by assigning the same value to all $z$ variables: i.e.,
\begin{align}
z^0_{hl} &= \zeta & \forall h,l \in H,\ \zeta \in\ ]0,1[ \label{eq:core_point}
\end{align}

\noindent
The Pareto subproblem for a trip $r$ then becomes 
\named{bus_pareto}%
    {Pareto}%
    {\max}%
    {u_o^r - u_d^r - \sum_{h,l \in H} z^0_{hl} v_{hl}}%
    {  u_i^r- u_j^r & \leq \tau_{ij} & \forall i,j \in D \\
    && u_h^r - u_l^r - v_{hl}^r & \leq \gamma_{hl} & \forall h, l \in H \\
    && u_o^r - u_d^r - \sum_{h,l \in H} \bar{z}_{hl} v_{hl} &= \sigma \label{eq:pareto-cut}}%
    {u_i^r \geq 0, v_{hl}^r \geq 0}

\noindent 
where $\sigma$ is the optimal objective value to the original
subproblem for trip $r$.

Our implementation also upates the core point, which can be seen as an
intensification procedure: Seldom used bus legs decay towards low
values while bus legs present in every solution are assigned a high
coefficient in Pareto solutions. The update rule is defined as follows:

\begin{equation}
z^{0 (k + 1)}_{hl} = \frac{z^{0 (k)}_{hl} + \bar{z}^{(k)}_{hl}}{2} \;\; \forall h,l \in H.
\end{equation}

\section{Experimental Results}
\label{section:results}

This section presents experimental results on the HSTPS problem. It
starts by specifying the exprimental setting. It continues by
justifying the implementation choices, including the Benders scheme,
the cut bundling strategy, and the core point updating rule. It also
compares the final algorithm with a standard MIP approach. Finally,
the section evaluates the impact of the algorithm on the real
case-study that motivated this work: The restructing of the Canberra
public transit system.

\subsection{Experimental Setting}

The results in Section \ref{sec:opti} already indicated that the
difference between the weekdays is minimal. As a result, the
experimental results use a single day and two sets of instances: A set
of small instances with a number of trips ranging from 100 to over
2,000 trips; and a set of large instances ranging from 1,000 to over
20,000 trips. The small instances are primarily used to eliminate
certain algorithmic options. The large instances are useful to
demonstrate scalability of the various advanced features. Unless
specified otherwise, the algorithms are used with the parameter
settings presented in Table \ref{tab:params}. Note that the time
penalty emulates the waiting time associated with any bus trip, i.e.,
it includes at least ten minutes of waiting time -- five minutes at
the first and last stops.

\begin{table}[t]
\centering
\begin{tabular}{*{6}{c}}
\hline
$c$ (\$/km) & $b$ (\$/km) & $S$ (s) & $n$ & $\alpha$ & Hubs \\
\midrule
1.96 & 4.5 & 30 & 32 & $10^{-3}$ &$H_{10}$  \\
\hline
\end{tabular}
\caption{The Default Values of the Parameters.}
\label{tab:params}
\end{table}

The LP/MIP solver used in the experiments in Gurobi v6.0
\citep{gurobi} with default parameters using the Dual Simplex and
without pre-processing. Gurobi was used to solve the MIP formulation,
the Benders master problem, and the sub-problems. The algorithms are
implemented in C++ and run as single threaded programs, forcing Gurobi
to use a single thread as well. The experiments were run on a cluster
with AMD Opteron 4184 CPUs and 64GB of RAM.

\subsection{Justification of the Benders Approach}

\subsubsection{Benders's Schemes}

\begin{figure}[th!]
    \centering
    \small
\begingroup
  \makeatletter
  \providecommand\color[2][]{%
    \GenericError{(gnuplot) \space\space\space\@spaces}{%
      Package color not loaded in conjunction with
      terminal option `colourtext'%
    }{See the gnuplot documentation for explanation.%
    }{Either use 'blacktext' in gnuplot or load the package
      color.sty in LaTeX.}%
    \renewcommand\color[2][]{}%
  }%
  \providecommand\includegraphics[2][]{%
    \GenericError{(gnuplot) \space\space\space\@spaces}{%
      Package graphicx or graphics not loaded%
    }{See the gnuplot documentation for explanation.%
    }{The gnuplot epslatex terminal needs graphicx.sty or graphics.sty.}%
    \renewcommand\includegraphics[2][]{}%
  }%
  \providecommand\rotatebox[2]{#2}%
  \@ifundefined{ifGPcolor}{%
    \newif\ifGPcolor
    \GPcolortrue
  }{}%
  \@ifundefined{ifGPblacktext}{%
    \newif\ifGPblacktext
    \GPblacktextfalse
  }{}%
  \let\gplgaddtomacro\g@addto@macro
  \gdef\gplbacktext{}%
  \gdef\gplfronttext{}%
  \makeatother
  \ifGPblacktext
    \def\colorrgb#1{}%
    \def\colorgray#1{}%
  \else
    \ifGPcolor
      \def\colorrgb#1{\color[rgb]{#1}}%
      \def\colorgray#1{\color[gray]{#1}}%
      \expandafter\def\csname LTw\endcsname{\color{white}}%
      \expandafter\def\csname LTb\endcsname{\color{black}}%
      \expandafter\def\csname LTa\endcsname{\color{black}}%
      \expandafter\def\csname LT0\endcsname{\color[rgb]{1,0,0}}%
      \expandafter\def\csname LT1\endcsname{\color[rgb]{0,1,0}}%
      \expandafter\def\csname LT2\endcsname{\color[rgb]{0,0,1}}%
      \expandafter\def\csname LT3\endcsname{\color[rgb]{1,0,1}}%
      \expandafter\def\csname LT4\endcsname{\color[rgb]{0,1,1}}%
      \expandafter\def\csname LT5\endcsname{\color[rgb]{1,1,0}}%
      \expandafter\def\csname LT6\endcsname{\color[rgb]{0,0,0}}%
      \expandafter\def\csname LT7\endcsname{\color[rgb]{1,0.3,0}}%
      \expandafter\def\csname LT8\endcsname{\color[rgb]{0.5,0.5,0.5}}%
    \else
      \def\colorrgb#1{\color{black}}%
      \def\colorgray#1{\color[gray]{#1}}%
      \expandafter\def\csname LTw\endcsname{\color{white}}%
      \expandafter\def\csname LTb\endcsname{\color{black}}%
      \expandafter\def\csname LTa\endcsname{\color{black}}%
      \expandafter\def\csname LT0\endcsname{\color{black}}%
      \expandafter\def\csname LT1\endcsname{\color{black}}%
      \expandafter\def\csname LT2\endcsname{\color{black}}%
      \expandafter\def\csname LT3\endcsname{\color{black}}%
      \expandafter\def\csname LT4\endcsname{\color{black}}%
      \expandafter\def\csname LT5\endcsname{\color{black}}%
      \expandafter\def\csname LT6\endcsname{\color{black}}%
      \expandafter\def\csname LT7\endcsname{\color{black}}%
      \expandafter\def\csname LT8\endcsname{\color{black}}%
    \fi
  \fi
    \setlength{\unitlength}{0.0500bp}%
    \ifx\gptboxheight\undefined%
      \newlength{\gptboxheight}%
      \newlength{\gptboxwidth}%
      \newsavebox{\gptboxtext}%
    \fi%
    \setlength{\fboxrule}{0.5pt}%
    \setlength{\fboxsep}{1pt}%
\begin{picture}(7200.00,5040.00)%
    \gplgaddtomacro\gplbacktext{%
      \colorrgb{0.50,0.50,0.50}%
      \put(1078,857){\makebox(0,0)[r]{\strut{}$0.01$}}%
      \colorrgb{0.50,0.50,0.50}%
      \put(1078,1510){\makebox(0,0)[r]{\strut{}$0.1$}}%
      \colorrgb{0.50,0.50,0.50}%
      \put(1078,2163){\makebox(0,0)[r]{\strut{}$1$}}%
      \colorrgb{0.50,0.50,0.50}%
      \put(1078,2816){\makebox(0,0)[r]{\strut{}$10$}}%
      \colorrgb{0.50,0.50,0.50}%
      \put(1078,3469){\makebox(0,0)[r]{\strut{}$100$}}%
      \colorrgb{0.50,0.50,0.50}%
      \put(1078,4122){\makebox(0,0)[r]{\strut{}$1000$}}%
      \colorrgb{0.50,0.50,0.50}%
      \put(1078,4775){\makebox(0,0)[r]{\strut{}$10000$}}%
      \colorrgb{0.50,0.50,0.50}%
      \put(1210,725){\rotatebox{-45}{\makebox(0,0)[l]{\strut{}100}}}%
      \colorrgb{0.50,0.50,0.50}%
      \put(1490,725){\rotatebox{-45}{\makebox(0,0)[l]{\strut{}200}}}%
      \colorrgb{0.50,0.50,0.50}%
      \put(1769,725){\rotatebox{-45}{\makebox(0,0)[l]{\strut{}300}}}%
      \colorrgb{0.50,0.50,0.50}%
      \put(2049,725){\rotatebox{-45}{\makebox(0,0)[l]{\strut{}400}}}%
      \colorrgb{0.50,0.50,0.50}%
      \put(2329,725){\rotatebox{-45}{\makebox(0,0)[l]{\strut{}500}}}%
      \colorrgb{0.50,0.50,0.50}%
      \put(2608,725){\rotatebox{-45}{\makebox(0,0)[l]{\strut{}600}}}%
      \colorrgb{0.50,0.50,0.50}%
      \put(2888,725){\rotatebox{-45}{\makebox(0,0)[l]{\strut{}700}}}%
      \colorrgb{0.50,0.50,0.50}%
      \put(3168,725){\rotatebox{-45}{\makebox(0,0)[l]{\strut{}800}}}%
      \colorrgb{0.50,0.50,0.50}%
      \put(3447,725){\rotatebox{-45}{\makebox(0,0)[l]{\strut{}900}}}%
      \colorrgb{0.50,0.50,0.50}%
      \put(3727,725){\rotatebox{-45}{\makebox(0,0)[l]{\strut{}1000}}}%
      \colorrgb{0.50,0.50,0.50}%
      \put(4007,725){\rotatebox{-45}{\makebox(0,0)[l]{\strut{}1100}}}%
      \colorrgb{0.50,0.50,0.50}%
      \put(4286,725){\rotatebox{-45}{\makebox(0,0)[l]{\strut{}1200}}}%
      \colorrgb{0.50,0.50,0.50}%
      \put(4566,725){\rotatebox{-45}{\makebox(0,0)[l]{\strut{}1300}}}%
      \colorrgb{0.50,0.50,0.50}%
      \put(4845,725){\rotatebox{-45}{\makebox(0,0)[l]{\strut{}1400}}}%
      \colorrgb{0.50,0.50,0.50}%
      \put(5125,725){\rotatebox{-45}{\makebox(0,0)[l]{\strut{}1500}}}%
      \colorrgb{0.50,0.50,0.50}%
      \put(5405,725){\rotatebox{-45}{\makebox(0,0)[l]{\strut{}1600}}}%
      \colorrgb{0.50,0.50,0.50}%
      \put(5684,725){\rotatebox{-45}{\makebox(0,0)[l]{\strut{}1700}}}%
      \colorrgb{0.50,0.50,0.50}%
      \put(5964,725){\rotatebox{-45}{\makebox(0,0)[l]{\strut{}1800}}}%
      \colorrgb{0.50,0.50,0.50}%
      \put(6244,725){\rotatebox{-45}{\makebox(0,0)[l]{\strut{}1900}}}%
      \colorrgb{0.50,0.50,0.50}%
      \put(6523,725){\rotatebox{-45}{\makebox(0,0)[l]{\strut{}2000}}}%
      \colorrgb{0.50,0.50,0.50}%
      \put(6803,725){\rotatebox{-45}{\makebox(0,0)[l]{\strut{}2100}}}%
    }%
    \gplgaddtomacro\gplfronttext{%
      \csname LTb\endcsname%
      \put(176,2816){\rotatebox{-270}{\makebox(0,0){\strut{}Time in s (log)}}}%
      \put(4006,154){\makebox(0,0){\strut{}Instance}}%
      \csname LTb\endcsname%
      \put(2197,4602){\makebox(0,0)[l]{\strut{}CP}}%
      \csname LTb\endcsname%
      \put(2197,4382){\makebox(0,0)[l]{\strut{}CR}}%
      \csname LTb\endcsname%
      \put(2197,4162){\makebox(0,0)[l]{\strut{}SP}}%
      \csname LTb\endcsname%
      \put(2197,3942){\makebox(0,0)[l]{\strut{}SR}}%
    }%
    \gplbacktext
    \put(0,0){\includegraphics{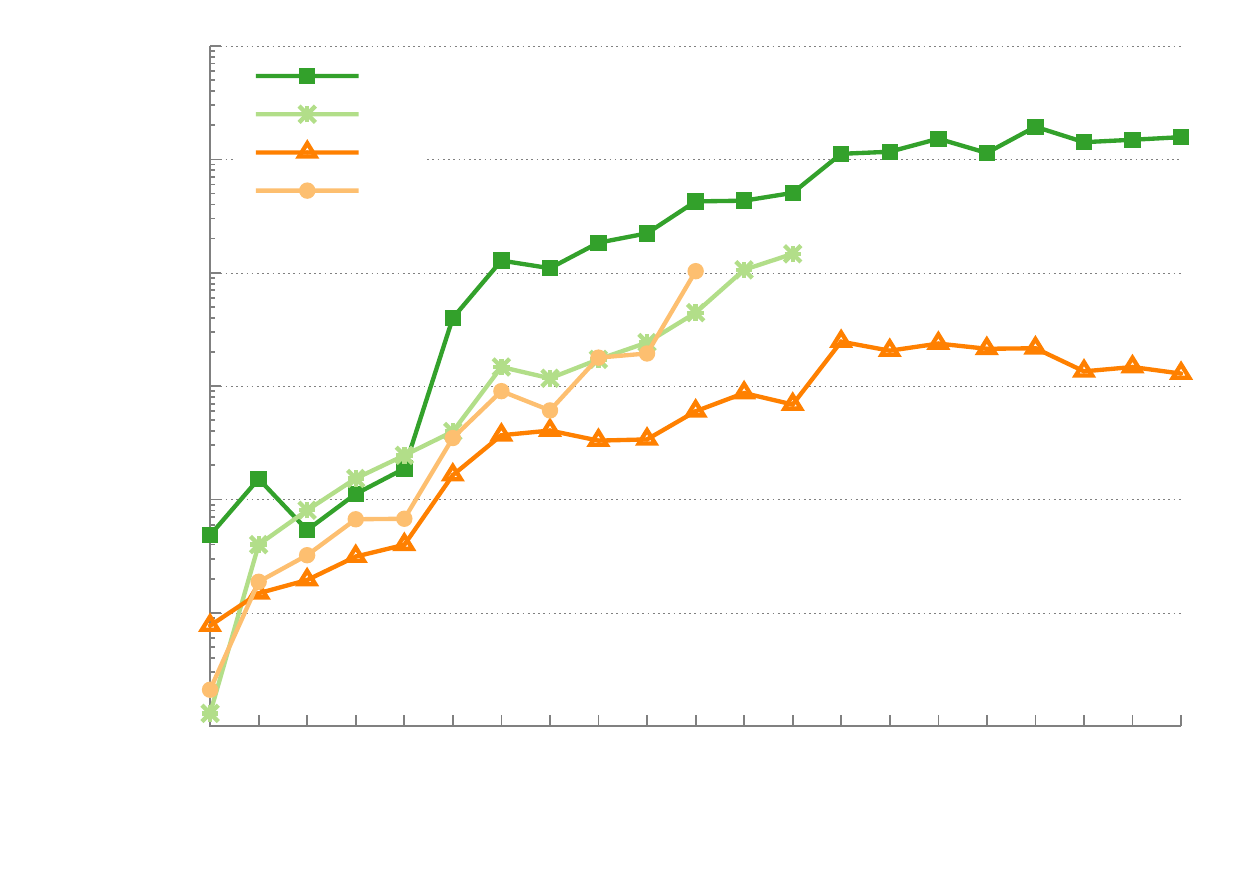}}%
    \gplfronttext
  \end{picture}%
\endgroup
    \caption{Comparison of the Benders Decomposition Schemes.}
    \label{fig:bend}
\end{figure}

Figure \ref{fig:bend} reports results of the following implementations of Benders decomposition:
\begin{enumerate}
\item [CR]: A single standard subproblem;
\item [CP]: A singe Pareto subproblem;
\item [SR]: Multiple independent standard subproblems;
\item [SP]: Multiple independent Pareto subproblems.
\end{enumerate}
The results are for the small set of instances and use the standard
parameters. The algorithms have a limit of 100 iterations: If the
Benders decomposition needs more than 100 iterations, it is most
likely exhibiting significant degeneracy.

The results in Figure \ref{fig:bend} clearly show the value of
Pareto-optimal cuts as the standard approach can only solve the lower
half of the instances before requiring more than a hundred
iterations. They also show that a single Pareto subproblem creates a
hard MIP problem and execution times become a significant issue.
Solving independent Pareto subproblems addresses both issues.

\subsubsection{Cut Aggregation Schemes}

\begin{figure}[th!]
    \centering
    \small
\begingroup
  \makeatletter
  \providecommand\color[2][]{%
    \GenericError{(gnuplot) \space\space\space\@spaces}{%
      Package color not loaded in conjunction with
      terminal option `colourtext'%
    }{See the gnuplot documentation for explanation.%
    }{Either use 'blacktext' in gnuplot or load the package
      color.sty in LaTeX.}%
    \renewcommand\color[2][]{}%
  }%
  \providecommand\includegraphics[2][]{%
    \GenericError{(gnuplot) \space\space\space\@spaces}{%
      Package graphicx or graphics not loaded%
    }{See the gnuplot documentation for explanation.%
    }{The gnuplot epslatex terminal needs graphicx.sty or graphics.sty.}%
    \renewcommand\includegraphics[2][]{}%
  }%
  \providecommand\rotatebox[2]{#2}%
  \@ifundefined{ifGPcolor}{%
    \newif\ifGPcolor
    \GPcolortrue
  }{}%
  \@ifundefined{ifGPblacktext}{%
    \newif\ifGPblacktext
    \GPblacktextfalse
  }{}%
  \let\gplgaddtomacro\g@addto@macro
  \gdef\gplbacktext{}%
  \gdef\gplfronttext{}%
  \makeatother
  \ifGPblacktext
    \def\colorrgb#1{}%
    \def\colorgray#1{}%
  \else
    \ifGPcolor
      \def\colorrgb#1{\color[rgb]{#1}}%
      \def\colorgray#1{\color[gray]{#1}}%
      \expandafter\def\csname LTw\endcsname{\color{white}}%
      \expandafter\def\csname LTb\endcsname{\color{black}}%
      \expandafter\def\csname LTa\endcsname{\color{black}}%
      \expandafter\def\csname LT0\endcsname{\color[rgb]{1,0,0}}%
      \expandafter\def\csname LT1\endcsname{\color[rgb]{0,1,0}}%
      \expandafter\def\csname LT2\endcsname{\color[rgb]{0,0,1}}%
      \expandafter\def\csname LT3\endcsname{\color[rgb]{1,0,1}}%
      \expandafter\def\csname LT4\endcsname{\color[rgb]{0,1,1}}%
      \expandafter\def\csname LT5\endcsname{\color[rgb]{1,1,0}}%
      \expandafter\def\csname LT6\endcsname{\color[rgb]{0,0,0}}%
      \expandafter\def\csname LT7\endcsname{\color[rgb]{1,0.3,0}}%
      \expandafter\def\csname LT8\endcsname{\color[rgb]{0.5,0.5,0.5}}%
    \else
      \def\colorrgb#1{\color{black}}%
      \def\colorgray#1{\color[gray]{#1}}%
      \expandafter\def\csname LTw\endcsname{\color{white}}%
      \expandafter\def\csname LTb\endcsname{\color{black}}%
      \expandafter\def\csname LTa\endcsname{\color{black}}%
      \expandafter\def\csname LT0\endcsname{\color{black}}%
      \expandafter\def\csname LT1\endcsname{\color{black}}%
      \expandafter\def\csname LT2\endcsname{\color{black}}%
      \expandafter\def\csname LT3\endcsname{\color{black}}%
      \expandafter\def\csname LT4\endcsname{\color{black}}%
      \expandafter\def\csname LT5\endcsname{\color{black}}%
      \expandafter\def\csname LT6\endcsname{\color{black}}%
      \expandafter\def\csname LT7\endcsname{\color{black}}%
      \expandafter\def\csname LT8\endcsname{\color{black}}%
    \fi
  \fi
    \setlength{\unitlength}{0.0500bp}%
    \ifx\gptboxheight\undefined%
      \newlength{\gptboxheight}%
      \newlength{\gptboxwidth}%
      \newsavebox{\gptboxtext}%
    \fi%
    \setlength{\fboxrule}{0.5pt}%
    \setlength{\fboxsep}{1pt}%
\begin{picture}(7200.00,5040.00)%
    \gplgaddtomacro\gplbacktext{%
      \colorrgb{0.50,0.50,0.50}%
      \put(1078,857){\makebox(0,0)[r]{\strut{}$0$}}%
      \colorrgb{0.50,0.50,0.50}%
      \put(1078,1347){\makebox(0,0)[r]{\strut{}$50$}}%
      \colorrgb{0.50,0.50,0.50}%
      \put(1078,1837){\makebox(0,0)[r]{\strut{}$100$}}%
      \colorrgb{0.50,0.50,0.50}%
      \put(1078,2326){\makebox(0,0)[r]{\strut{}$150$}}%
      \colorrgb{0.50,0.50,0.50}%
      \put(1078,2816){\makebox(0,0)[r]{\strut{}$200$}}%
      \colorrgb{0.50,0.50,0.50}%
      \put(1078,3306){\makebox(0,0)[r]{\strut{}$250$}}%
      \colorrgb{0.50,0.50,0.50}%
      \put(1078,3796){\makebox(0,0)[r]{\strut{}$300$}}%
      \colorrgb{0.50,0.50,0.50}%
      \put(1078,4285){\makebox(0,0)[r]{\strut{}$350$}}%
      \colorrgb{0.50,0.50,0.50}%
      \put(1078,4775){\makebox(0,0)[r]{\strut{}$400$}}%
      \colorrgb{0.50,0.50,0.50}%
      \put(1210,725){\rotatebox{-45}{\makebox(0,0)[l]{\strut{}100}}}%
      \colorrgb{0.50,0.50,0.50}%
      \put(1490,725){\rotatebox{-45}{\makebox(0,0)[l]{\strut{}200}}}%
      \colorrgb{0.50,0.50,0.50}%
      \put(1769,725){\rotatebox{-45}{\makebox(0,0)[l]{\strut{}300}}}%
      \colorrgb{0.50,0.50,0.50}%
      \put(2049,725){\rotatebox{-45}{\makebox(0,0)[l]{\strut{}400}}}%
      \colorrgb{0.50,0.50,0.50}%
      \put(2329,725){\rotatebox{-45}{\makebox(0,0)[l]{\strut{}500}}}%
      \colorrgb{0.50,0.50,0.50}%
      \put(2608,725){\rotatebox{-45}{\makebox(0,0)[l]{\strut{}600}}}%
      \colorrgb{0.50,0.50,0.50}%
      \put(2888,725){\rotatebox{-45}{\makebox(0,0)[l]{\strut{}700}}}%
      \colorrgb{0.50,0.50,0.50}%
      \put(3168,725){\rotatebox{-45}{\makebox(0,0)[l]{\strut{}800}}}%
      \colorrgb{0.50,0.50,0.50}%
      \put(3447,725){\rotatebox{-45}{\makebox(0,0)[l]{\strut{}900}}}%
      \colorrgb{0.50,0.50,0.50}%
      \put(3727,725){\rotatebox{-45}{\makebox(0,0)[l]{\strut{}1000}}}%
      \colorrgb{0.50,0.50,0.50}%
      \put(4007,725){\rotatebox{-45}{\makebox(0,0)[l]{\strut{}1100}}}%
      \colorrgb{0.50,0.50,0.50}%
      \put(4286,725){\rotatebox{-45}{\makebox(0,0)[l]{\strut{}1200}}}%
      \colorrgb{0.50,0.50,0.50}%
      \put(4566,725){\rotatebox{-45}{\makebox(0,0)[l]{\strut{}1300}}}%
      \colorrgb{0.50,0.50,0.50}%
      \put(4845,725){\rotatebox{-45}{\makebox(0,0)[l]{\strut{}1400}}}%
      \colorrgb{0.50,0.50,0.50}%
      \put(5125,725){\rotatebox{-45}{\makebox(0,0)[l]{\strut{}1500}}}%
      \colorrgb{0.50,0.50,0.50}%
      \put(5405,725){\rotatebox{-45}{\makebox(0,0)[l]{\strut{}1600}}}%
      \colorrgb{0.50,0.50,0.50}%
      \put(5684,725){\rotatebox{-45}{\makebox(0,0)[l]{\strut{}1700}}}%
      \colorrgb{0.50,0.50,0.50}%
      \put(5964,725){\rotatebox{-45}{\makebox(0,0)[l]{\strut{}1800}}}%
      \colorrgb{0.50,0.50,0.50}%
      \put(6244,725){\rotatebox{-45}{\makebox(0,0)[l]{\strut{}1900}}}%
      \colorrgb{0.50,0.50,0.50}%
      \put(6523,725){\rotatebox{-45}{\makebox(0,0)[l]{\strut{}2000}}}%
      \colorrgb{0.50,0.50,0.50}%
      \put(6803,725){\rotatebox{-45}{\makebox(0,0)[l]{\strut{}2100}}}%
    }%
    \gplgaddtomacro\gplfronttext{%
      \csname LTb\endcsname%
      \put(176,2816){\rotatebox{-270}{\makebox(0,0){\strut{}Time in s}}}%
      \put(4006,154){\makebox(0,0){\strut{}Instance}}%
      \csname LTb\endcsname%
      \put(2197,4602){\makebox(0,0)[l]{\strut{}Hubs}}%
      \csname LTb\endcsname%
      \put(2197,4382){\makebox(0,0)[l]{\strut{}Legs}}%
      \csname LTb\endcsname%
      \put(2197,4162){\makebox(0,0)[l]{\strut{}Multi}}%
      \csname LTb\endcsname%
      \put(2197,3942){\makebox(0,0)[l]{\strut{}One}}%
      \csname LTb\endcsname%
      \put(2197,3722){\makebox(0,0)[l]{\strut{}Origin}}%
    }%
    \gplbacktext
    \put(0,0){\includegraphics{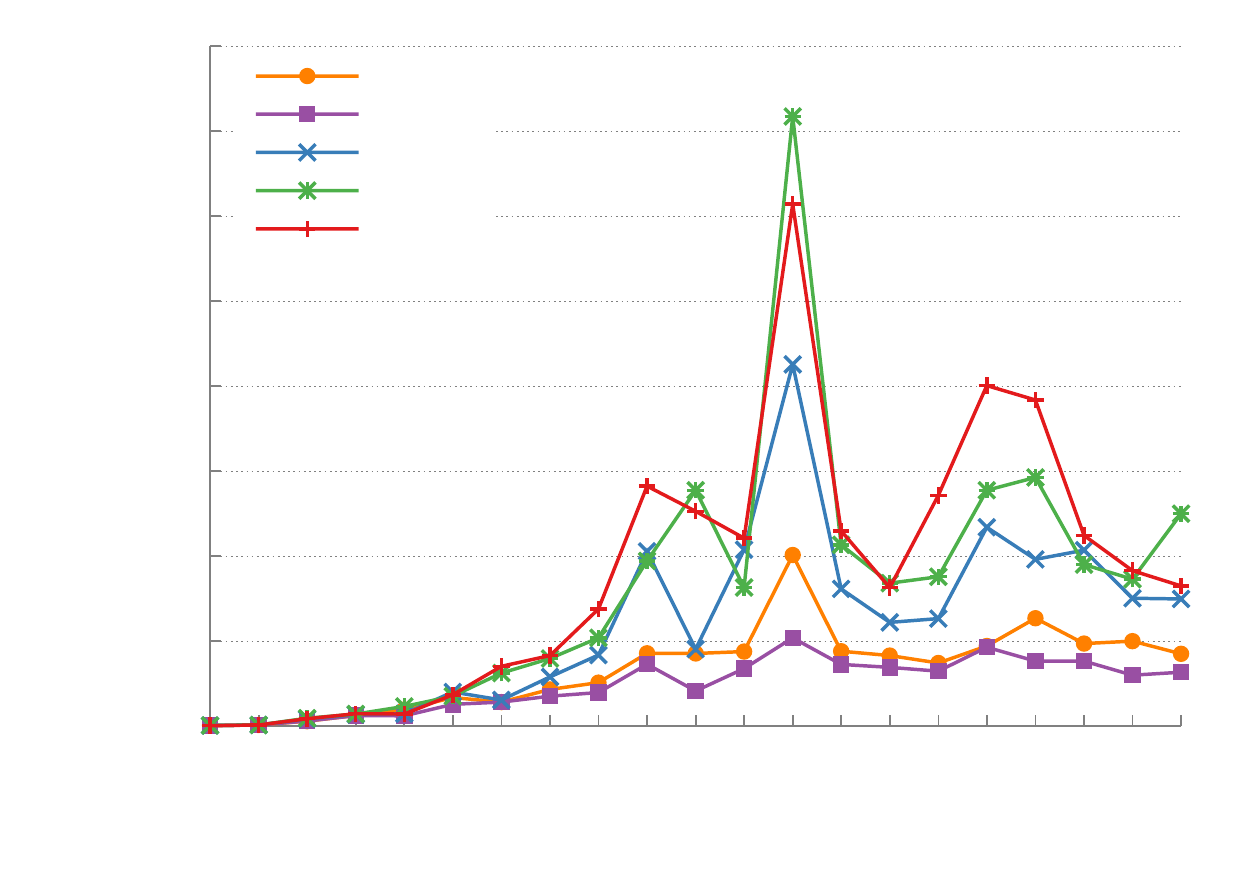}}%
    \gplfronttext
  \end{picture}%
\endgroup
    \caption{Comparison of Aggregation Schemes on Small Instances}
    \label{fig:agg_100}
\end{figure}

\begin{figure}[th!]
    \centering
    \small
\begingroup
  \makeatletter
  \providecommand\color[2][]{%
    \GenericError{(gnuplot) \space\space\space\@spaces}{%
      Package color not loaded in conjunction with
      terminal option `colourtext'%
    }{See the gnuplot documentation for explanation.%
    }{Either use 'blacktext' in gnuplot or load the package
      color.sty in LaTeX.}%
    \renewcommand\color[2][]{}%
  }%
  \providecommand\includegraphics[2][]{%
    \GenericError{(gnuplot) \space\space\space\@spaces}{%
      Package graphicx or graphics not loaded%
    }{See the gnuplot documentation for explanation.%
    }{The gnuplot epslatex terminal needs graphicx.sty or graphics.sty.}%
    \renewcommand\includegraphics[2][]{}%
  }%
  \providecommand\rotatebox[2]{#2}%
  \@ifundefined{ifGPcolor}{%
    \newif\ifGPcolor
    \GPcolortrue
  }{}%
  \@ifundefined{ifGPblacktext}{%
    \newif\ifGPblacktext
    \GPblacktextfalse
  }{}%
  \let\gplgaddtomacro\g@addto@macro
  \gdef\gplbacktext{}%
  \gdef\gplfronttext{}%
  \makeatother
  \ifGPblacktext
    \def\colorrgb#1{}%
    \def\colorgray#1{}%
  \else
    \ifGPcolor
      \def\colorrgb#1{\color[rgb]{#1}}%
      \def\colorgray#1{\color[gray]{#1}}%
      \expandafter\def\csname LTw\endcsname{\color{white}}%
      \expandafter\def\csname LTb\endcsname{\color{black}}%
      \expandafter\def\csname LTa\endcsname{\color{black}}%
      \expandafter\def\csname LT0\endcsname{\color[rgb]{1,0,0}}%
      \expandafter\def\csname LT1\endcsname{\color[rgb]{0,1,0}}%
      \expandafter\def\csname LT2\endcsname{\color[rgb]{0,0,1}}%
      \expandafter\def\csname LT3\endcsname{\color[rgb]{1,0,1}}%
      \expandafter\def\csname LT4\endcsname{\color[rgb]{0,1,1}}%
      \expandafter\def\csname LT5\endcsname{\color[rgb]{1,1,0}}%
      \expandafter\def\csname LT6\endcsname{\color[rgb]{0,0,0}}%
      \expandafter\def\csname LT7\endcsname{\color[rgb]{1,0.3,0}}%
      \expandafter\def\csname LT8\endcsname{\color[rgb]{0.5,0.5,0.5}}%
    \else
      \def\colorrgb#1{\color{black}}%
      \def\colorgray#1{\color[gray]{#1}}%
      \expandafter\def\csname LTw\endcsname{\color{white}}%
      \expandafter\def\csname LTb\endcsname{\color{black}}%
      \expandafter\def\csname LTa\endcsname{\color{black}}%
      \expandafter\def\csname LT0\endcsname{\color{black}}%
      \expandafter\def\csname LT1\endcsname{\color{black}}%
      \expandafter\def\csname LT2\endcsname{\color{black}}%
      \expandafter\def\csname LT3\endcsname{\color{black}}%
      \expandafter\def\csname LT4\endcsname{\color{black}}%
      \expandafter\def\csname LT5\endcsname{\color{black}}%
      \expandafter\def\csname LT6\endcsname{\color{black}}%
      \expandafter\def\csname LT7\endcsname{\color{black}}%
      \expandafter\def\csname LT8\endcsname{\color{black}}%
    \fi
  \fi
    \setlength{\unitlength}{0.0500bp}%
    \ifx\gptboxheight\undefined%
      \newlength{\gptboxheight}%
      \newlength{\gptboxwidth}%
      \newsavebox{\gptboxtext}%
    \fi%
    \setlength{\fboxrule}{0.5pt}%
    \setlength{\fboxsep}{1pt}%
\begin{picture}(7200.00,5040.00)%
    \gplgaddtomacro\gplbacktext{%
      \colorrgb{0.50,0.50,0.50}%
      \put(814,950){\makebox(0,0)[r]{\strut{}$0$}}%
      \colorrgb{0.50,0.50,0.50}%
      \put(814,1428){\makebox(0,0)[r]{\strut{}$100$}}%
      \colorrgb{0.50,0.50,0.50}%
      \put(814,1906){\makebox(0,0)[r]{\strut{}$200$}}%
      \colorrgb{0.50,0.50,0.50}%
      \put(814,2384){\makebox(0,0)[r]{\strut{}$300$}}%
      \colorrgb{0.50,0.50,0.50}%
      \put(814,2863){\makebox(0,0)[r]{\strut{}$400$}}%
      \colorrgb{0.50,0.50,0.50}%
      \put(814,3341){\makebox(0,0)[r]{\strut{}$500$}}%
      \colorrgb{0.50,0.50,0.50}%
      \put(814,3819){\makebox(0,0)[r]{\strut{}$600$}}%
      \colorrgb{0.50,0.50,0.50}%
      \put(814,4297){\makebox(0,0)[r]{\strut{}$700$}}%
      \colorrgb{0.50,0.50,0.50}%
      \put(814,4775){\makebox(0,0)[r]{\strut{}$800$}}%
      \colorrgb{0.50,0.50,0.50}%
      \put(946,818){\rotatebox{-45}{\makebox(0,0)[l]{\strut{}1000}}}%
      \colorrgb{0.50,0.50,0.50}%
      \put(1229,818){\rotatebox{-45}{\makebox(0,0)[l]{\strut{}2000}}}%
      \colorrgb{0.50,0.50,0.50}%
      \put(1512,818){\rotatebox{-45}{\makebox(0,0)[l]{\strut{}3000}}}%
      \colorrgb{0.50,0.50,0.50}%
      \put(1795,818){\rotatebox{-45}{\makebox(0,0)[l]{\strut{}4000}}}%
      \colorrgb{0.50,0.50,0.50}%
      \put(2078,818){\rotatebox{-45}{\makebox(0,0)[l]{\strut{}5000}}}%
      \colorrgb{0.50,0.50,0.50}%
      \put(2361,818){\rotatebox{-45}{\makebox(0,0)[l]{\strut{}6000}}}%
      \colorrgb{0.50,0.50,0.50}%
      \put(2644,818){\rotatebox{-45}{\makebox(0,0)[l]{\strut{}7000}}}%
      \colorrgb{0.50,0.50,0.50}%
      \put(2926,818){\rotatebox{-45}{\makebox(0,0)[l]{\strut{}8000}}}%
      \colorrgb{0.50,0.50,0.50}%
      \put(3209,818){\rotatebox{-45}{\makebox(0,0)[l]{\strut{}9000}}}%
      \colorrgb{0.50,0.50,0.50}%
      \put(3492,818){\rotatebox{-45}{\makebox(0,0)[l]{\strut{}10000}}}%
      \colorrgb{0.50,0.50,0.50}%
      \put(3775,818){\rotatebox{-45}{\makebox(0,0)[l]{\strut{}11000}}}%
      \colorrgb{0.50,0.50,0.50}%
      \put(4058,818){\rotatebox{-45}{\makebox(0,0)[l]{\strut{}12000}}}%
      \colorrgb{0.50,0.50,0.50}%
      \put(4341,818){\rotatebox{-45}{\makebox(0,0)[l]{\strut{}13000}}}%
      \colorrgb{0.50,0.50,0.50}%
      \put(4624,818){\rotatebox{-45}{\makebox(0,0)[l]{\strut{}14000}}}%
      \colorrgb{0.50,0.50,0.50}%
      \put(4907,818){\rotatebox{-45}{\makebox(0,0)[l]{\strut{}15000}}}%
      \colorrgb{0.50,0.50,0.50}%
      \put(5190,818){\rotatebox{-45}{\makebox(0,0)[l]{\strut{}16000}}}%
      \colorrgb{0.50,0.50,0.50}%
      \put(5473,818){\rotatebox{-45}{\makebox(0,0)[l]{\strut{}17000}}}%
      \colorrgb{0.50,0.50,0.50}%
      \put(5756,818){\rotatebox{-45}{\makebox(0,0)[l]{\strut{}18000}}}%
      \colorrgb{0.50,0.50,0.50}%
      \put(6039,818){\rotatebox{-45}{\makebox(0,0)[l]{\strut{}19000}}}%
      \colorrgb{0.50,0.50,0.50}%
      \put(6321,818){\rotatebox{-45}{\makebox(0,0)[l]{\strut{}20000}}}%
      \colorrgb{0.50,0.50,0.50}%
      \put(6604,818){\rotatebox{-45}{\makebox(0,0)[l]{\strut{}21000}}}%
    }%
    \gplgaddtomacro\gplfronttext{%
      \csname LTb\endcsname%
      \put(176,2862){\rotatebox{-270}{\makebox(0,0){\strut{}Time in s}}}%
      \put(3803,154){\makebox(0,0){\strut{}Instance}}%
      \csname LTb\endcsname%
      \put(1933,4602){\makebox(0,0)[l]{\strut{}Hubs}}%
      \csname LTb\endcsname%
      \put(1933,4382){\makebox(0,0)[l]{\strut{}Legs}}%
      \csname LTb\endcsname%
      \put(1933,4162){\makebox(0,0)[l]{\strut{}One}}%
    }%
    \gplbacktext
    \put(0,0){\includegraphics{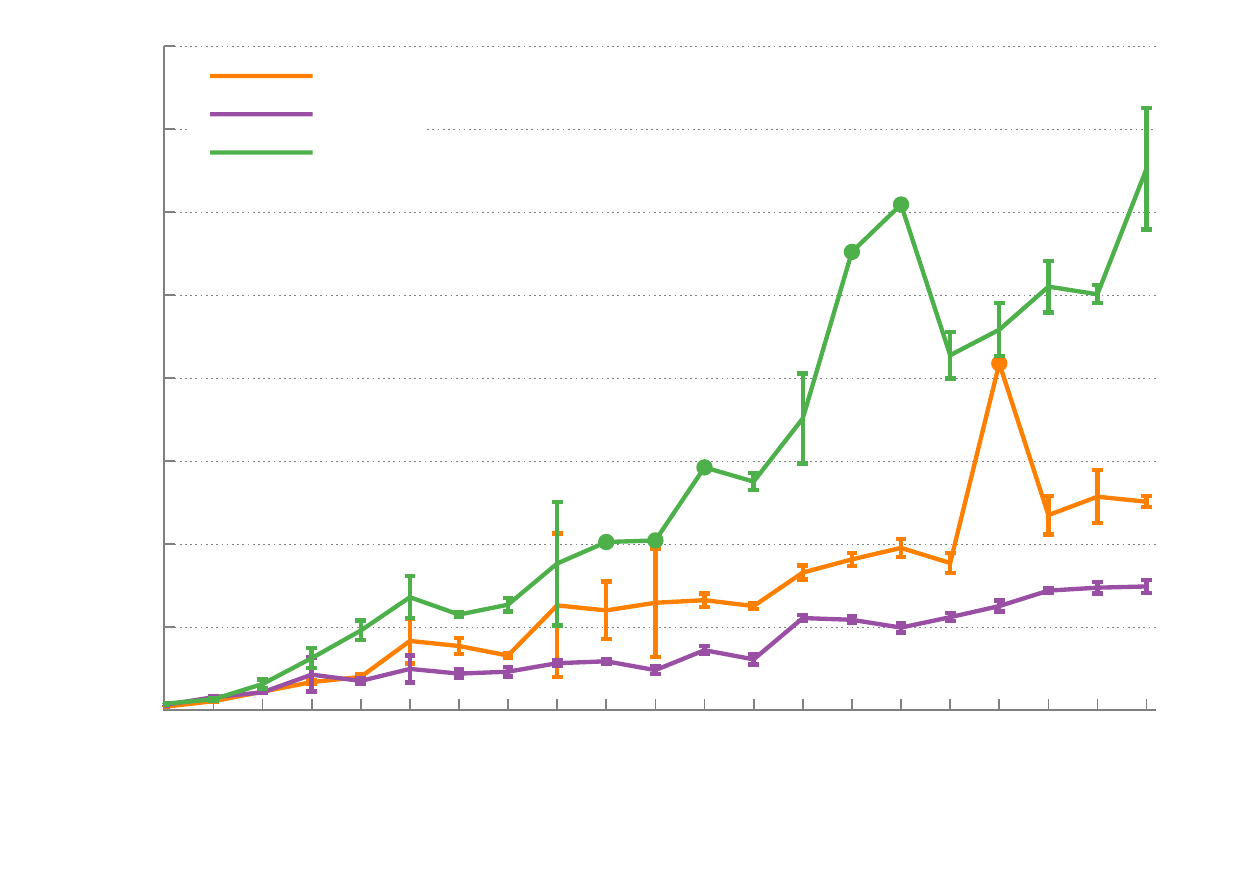}}%
    \gplfronttext
  \end{picture}%
\endgroup
    \caption{Comparing Cut Bundling Schemes.}
    \label{fig:errors}
\end{figure}

\begin{figure}[th!]
    \centering
    \small
\begingroup
  \makeatletter
  \providecommand\color[2][]{%
    \GenericError{(gnuplot) \space\space\space\@spaces}{%
      Package color not loaded in conjunction with
      terminal option `colourtext'%
    }{See the gnuplot documentation for explanation.%
    }{Either use 'blacktext' in gnuplot or load the package
      color.sty in LaTeX.}%
    \renewcommand\color[2][]{}%
  }%
  \providecommand\includegraphics[2][]{%
    \GenericError{(gnuplot) \space\space\space\@spaces}{%
      Package graphicx or graphics not loaded%
    }{See the gnuplot documentation for explanation.%
    }{The gnuplot epslatex terminal needs graphicx.sty or graphics.sty.}%
    \renewcommand\includegraphics[2][]{}%
  }%
  \providecommand\rotatebox[2]{#2}%
  \@ifundefined{ifGPcolor}{%
    \newif\ifGPcolor
    \GPcolortrue
  }{}%
  \@ifundefined{ifGPblacktext}{%
    \newif\ifGPblacktext
    \GPblacktextfalse
  }{}%
  \let\gplgaddtomacro\g@addto@macro
  \gdef\gplbacktext{}%
  \gdef\gplfronttext{}%
  \makeatother
  \ifGPblacktext
    \def\colorrgb#1{}%
    \def\colorgray#1{}%
  \else
    \ifGPcolor
      \def\colorrgb#1{\color[rgb]{#1}}%
      \def\colorgray#1{\color[gray]{#1}}%
      \expandafter\def\csname LTw\endcsname{\color{white}}%
      \expandafter\def\csname LTb\endcsname{\color{black}}%
      \expandafter\def\csname LTa\endcsname{\color{black}}%
      \expandafter\def\csname LT0\endcsname{\color[rgb]{1,0,0}}%
      \expandafter\def\csname LT1\endcsname{\color[rgb]{0,1,0}}%
      \expandafter\def\csname LT2\endcsname{\color[rgb]{0,0,1}}%
      \expandafter\def\csname LT3\endcsname{\color[rgb]{1,0,1}}%
      \expandafter\def\csname LT4\endcsname{\color[rgb]{0,1,1}}%
      \expandafter\def\csname LT5\endcsname{\color[rgb]{1,1,0}}%
      \expandafter\def\csname LT6\endcsname{\color[rgb]{0,0,0}}%
      \expandafter\def\csname LT7\endcsname{\color[rgb]{1,0.3,0}}%
      \expandafter\def\csname LT8\endcsname{\color[rgb]{0.5,0.5,0.5}}%
    \else
      \def\colorrgb#1{\color{black}}%
      \def\colorgray#1{\color[gray]{#1}}%
      \expandafter\def\csname LTw\endcsname{\color{white}}%
      \expandafter\def\csname LTb\endcsname{\color{black}}%
      \expandafter\def\csname LTa\endcsname{\color{black}}%
      \expandafter\def\csname LT0\endcsname{\color{black}}%
      \expandafter\def\csname LT1\endcsname{\color{black}}%
      \expandafter\def\csname LT2\endcsname{\color{black}}%
      \expandafter\def\csname LT3\endcsname{\color{black}}%
      \expandafter\def\csname LT4\endcsname{\color{black}}%
      \expandafter\def\csname LT5\endcsname{\color{black}}%
      \expandafter\def\csname LT6\endcsname{\color{black}}%
      \expandafter\def\csname LT7\endcsname{\color{black}}%
      \expandafter\def\csname LT8\endcsname{\color{black}}%
    \fi
  \fi
    \setlength{\unitlength}{0.0500bp}%
    \ifx\gptboxheight\undefined%
      \newlength{\gptboxheight}%
      \newlength{\gptboxwidth}%
      \newsavebox{\gptboxtext}%
    \fi%
    \setlength{\fboxrule}{0.5pt}%
    \setlength{\fboxsep}{1pt}%
\begin{picture}(7200.00,5040.00)%
    \gplgaddtomacro\gplbacktext{%
      \colorrgb{0.50,0.50,0.50}%
      \put(682,3206){\makebox(0,0)[r]{\strut{}$0$}}%
      \colorrgb{0.50,0.50,0.50}%
      \put(682,3380){\makebox(0,0)[r]{\strut{}$5$}}%
      \colorrgb{0.50,0.50,0.50}%
      \put(682,3555){\makebox(0,0)[r]{\strut{}$10$}}%
      \colorrgb{0.50,0.50,0.50}%
      \put(682,3729){\makebox(0,0)[r]{\strut{}$15$}}%
      \colorrgb{0.50,0.50,0.50}%
      \put(682,3903){\makebox(0,0)[r]{\strut{}$20$}}%
      \colorrgb{0.50,0.50,0.50}%
      \put(682,4078){\makebox(0,0)[r]{\strut{}$25$}}%
      \colorrgb{0.50,0.50,0.50}%
      \put(682,4252){\makebox(0,0)[r]{\strut{}$30$}}%
      \colorrgb{0.50,0.50,0.50}%
      \put(682,4426){\makebox(0,0)[r]{\strut{}$35$}}%
      \colorrgb{0.50,0.50,0.50}%
      \put(682,4601){\makebox(0,0)[r]{\strut{}$40$}}%
      \colorrgb{0.50,0.50,0.50}%
      \put(682,4775){\makebox(0,0)[r]{\strut{}$45$}}%
      \colorrgb{0.50,0.50,0.50}%
      \put(1113,3074){\rotatebox{-45}{\makebox(0,0)[l]{\strut{}1000}}}%
      \colorrgb{0.50,0.50,0.50}%
      \put(1712,3074){\rotatebox{-45}{\makebox(0,0)[l]{\strut{}2000}}}%
      \colorrgb{0.50,0.50,0.50}%
      \put(2311,3074){\rotatebox{-45}{\makebox(0,0)[l]{\strut{}3000}}}%
      \colorrgb{0.50,0.50,0.50}%
      \put(2910,3074){\rotatebox{-45}{\makebox(0,0)[l]{\strut{}4000}}}%
      \colorrgb{0.50,0.50,0.50}%
      \put(3509,3074){\rotatebox{-45}{\makebox(0,0)[l]{\strut{}5000}}}%
      \colorrgb{0.50,0.50,0.50}%
      \put(4108,3074){\rotatebox{-45}{\makebox(0,0)[l]{\strut{}6000}}}%
      \colorrgb{0.50,0.50,0.50}%
      \put(4707,3074){\rotatebox{-45}{\makebox(0,0)[l]{\strut{}7000}}}%
      \colorrgb{0.50,0.50,0.50}%
      \put(5306,3074){\rotatebox{-45}{\makebox(0,0)[l]{\strut{}8000}}}%
      \colorrgb{0.50,0.50,0.50}%
      \put(5905,3074){\rotatebox{-45}{\makebox(0,0)[l]{\strut{}9000}}}%
      \colorrgb{0.50,0.50,0.50}%
      \put(6504,3074){\rotatebox{-45}{\makebox(0,0)[l]{\strut{}10000}}}%
    }%
    \gplgaddtomacro\gplfronttext{%
      \csname LTb\endcsname%
      \put(176,3990){\rotatebox{-270}{\makebox(0,0){\strut{}Iterations}}}%
      \csname LTb\endcsname%
      \put(1801,4602){\makebox(0,0)[l]{\strut{}Hubs}}%
      \csname LTb\endcsname%
      \put(1801,4382){\makebox(0,0)[l]{\strut{}Legs}}%
      \csname LTb\endcsname%
      \put(1801,4162){\makebox(0,0)[l]{\strut{}Multi}}%
      \csname LTb\endcsname%
      \put(1801,3942){\makebox(0,0)[l]{\strut{}One}}%
      \csname LTb\endcsname%
      \put(1801,3722){\makebox(0,0)[l]{\strut{}Origin}}%
    }%
    \gplgaddtomacro\gplbacktext{%
      \colorrgb{0.50,0.50,0.50}%
      \put(682,950){\makebox(0,0)[r]{\strut{}$0$}}%
      \colorrgb{0.50,0.50,0.50}%
      \put(682,1095){\makebox(0,0)[r]{\strut{}$10$}}%
      \colorrgb{0.50,0.50,0.50}%
      \put(682,1240){\makebox(0,0)[r]{\strut{}$20$}}%
      \colorrgb{0.50,0.50,0.50}%
      \put(682,1385){\makebox(0,0)[r]{\strut{}$30$}}%
      \colorrgb{0.50,0.50,0.50}%
      \put(682,1530){\makebox(0,0)[r]{\strut{}$40$}}%
      \colorrgb{0.50,0.50,0.50}%
      \put(682,1676){\makebox(0,0)[r]{\strut{}$50$}}%
      \colorrgb{0.50,0.50,0.50}%
      \put(682,1821){\makebox(0,0)[r]{\strut{}$60$}}%
      \colorrgb{0.50,0.50,0.50}%
      \put(682,1966){\makebox(0,0)[r]{\strut{}$70$}}%
      \colorrgb{0.50,0.50,0.50}%
      \put(682,2111){\makebox(0,0)[r]{\strut{}$80$}}%
      \colorrgb{0.50,0.50,0.50}%
      \put(682,2256){\makebox(0,0)[r]{\strut{}$90$}}%
      \colorrgb{0.50,0.50,0.50}%
      \put(1086,818){\rotatebox{-45}{\makebox(0,0)[l]{\strut{}11000}}}%
      \colorrgb{0.50,0.50,0.50}%
      \put(1631,818){\rotatebox{-45}{\makebox(0,0)[l]{\strut{}12000}}}%
      \colorrgb{0.50,0.50,0.50}%
      \put(2175,818){\rotatebox{-45}{\makebox(0,0)[l]{\strut{}13000}}}%
      \colorrgb{0.50,0.50,0.50}%
      \put(2720,818){\rotatebox{-45}{\makebox(0,0)[l]{\strut{}14000}}}%
      \colorrgb{0.50,0.50,0.50}%
      \put(3264,818){\rotatebox{-45}{\makebox(0,0)[l]{\strut{}15000}}}%
      \colorrgb{0.50,0.50,0.50}%
      \put(3809,818){\rotatebox{-45}{\makebox(0,0)[l]{\strut{}16000}}}%
      \colorrgb{0.50,0.50,0.50}%
      \put(4353,818){\rotatebox{-45}{\makebox(0,0)[l]{\strut{}17000}}}%
      \colorrgb{0.50,0.50,0.50}%
      \put(4897,818){\rotatebox{-45}{\makebox(0,0)[l]{\strut{}18000}}}%
      \colorrgb{0.50,0.50,0.50}%
      \put(5442,818){\rotatebox{-45}{\makebox(0,0)[l]{\strut{}19000}}}%
      \colorrgb{0.50,0.50,0.50}%
      \put(5986,818){\rotatebox{-45}{\makebox(0,0)[l]{\strut{}20000}}}%
      \colorrgb{0.50,0.50,0.50}%
      \put(6531,818){\rotatebox{-45}{\makebox(0,0)[l]{\strut{}21000}}}%
    }%
    \gplgaddtomacro\gplfronttext{%
      \csname LTb\endcsname%
      \put(176,1603){\rotatebox{-270}{\makebox(0,0){\strut{}Iterations}}}%
      \put(3808,154){\makebox(0,0){\strut{}Instance}}%
    }%
    \gplbacktext
    \put(0,0){\includegraphics{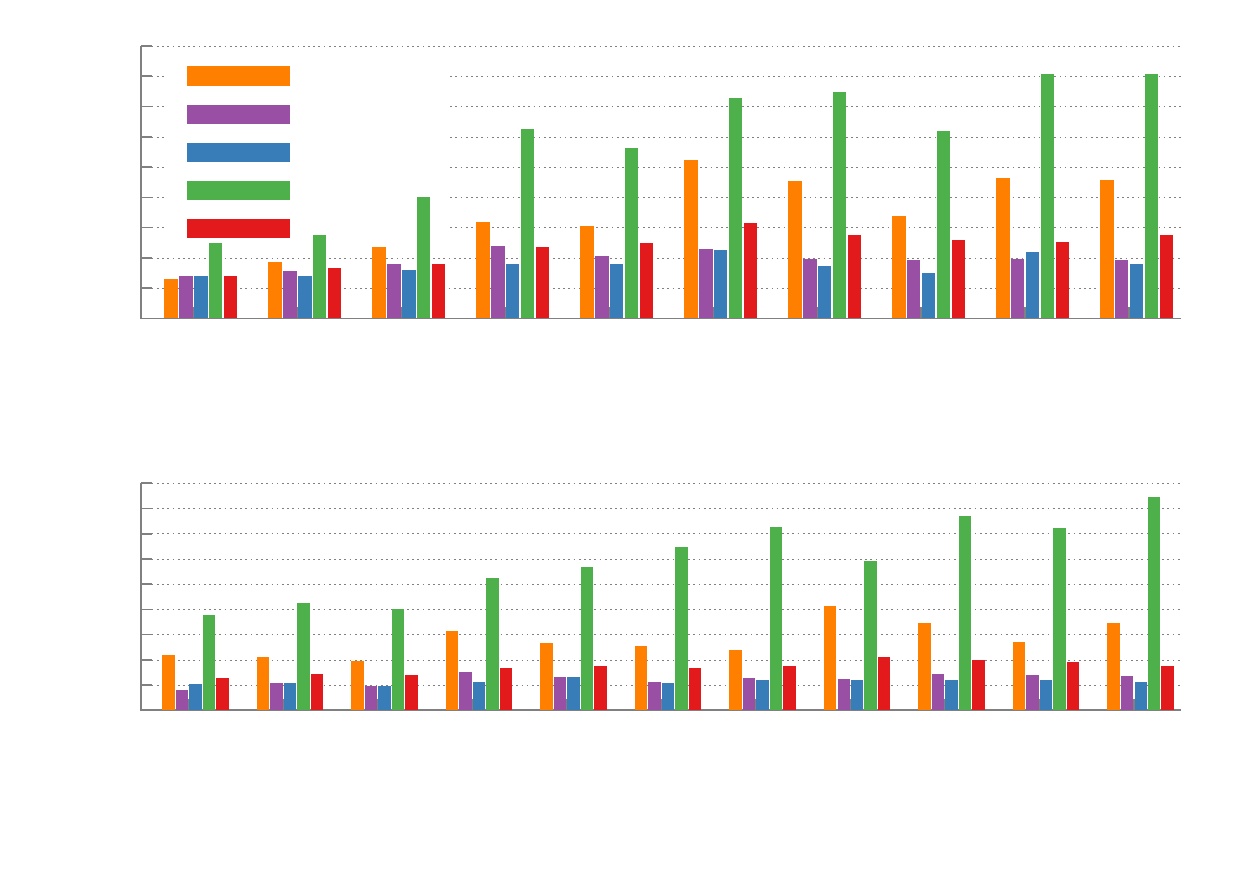}}%
    \gplfronttext
  \end{picture}%
\endgroup
    \caption{Number of Iterations for each Bundling Schemes on Larges Instances.}
    \label{fig:iterations}
\end{figure}

\begin{table}[th!]
\centering
\begin{tabular}{r *{4}{| c c}}
    \multirow{2}{*}{Instance} & \multicolumn{2}{c}{Hubs} & \multicolumn{2}{| c}{Legs} & \multicolumn{2}{| c}{Origin} & \multicolumn{2}{| c}{Multi} \\
    & $H_{10}$ & $H_{20}$ & $H_{10}$ & $H_{20}$ & $H_{10}$ & $H_{20}$ & $H_{10}$ & $H_{20}$ \\
    \hline
1000 & 10 & 20 & 83 & 234 & 395 & 491 & 658 & 811 \\
2000 & 10 & 20 & 90 & 280 & 607 & 716 & 1290 & 1558 \\
3000 & 10 & 20 & 89 & 309 & 830 & 935 & 2002 & 2400 \\
4000 & 10 & 20 & 89 & 322 & 962 & 1104 & 2659 & 3206 \\
5000 & 10 & 20 & 90 & 328 & 1109 & 1257 & 3364 & 4036 \\
6000 & 10 & 20 & 90 & 346 & 1200 & 1357 & 4036 & 4851 \\
7000 & 10 & 20 & 90 & 339 & 1262 & 1427 & 4613 & 5568 \\
8000 & 10 & 20 & 90 & 351 & 1360 & 1513 & 5315 & 6409 \\
9000 & 10 & 20 & 90 & 354 & 1387 & 1556 & 6008 & 7182 \\
10000 & 10 & 20 & 90 & 353 & 1485 & 1667 & 6741 & 8084 \\
11000 & 10 & 20 & 90 & 357 & 1528 & 1712 & 7329 & 8863 \\
12000 & 10 & 20 & 90 & 362 & 1582 & 1762 & 8010 & 9623 \\
13000 & 10 & 20 & 90 & 360 & 1598 & 1793 & 8678 & 10447 \\
14000 & 10 & 20 & 90 & 364 & 1644 & 1846 & 9290 & 11230 \\
15000 & 10 & 20 & 90 & 359 & 1680 & 1872 & 9943 & 12000 \\
16000 & 10 & 20 & 90 & 364 & 1721 & 1918 & 10659 & 12838 \\
17000 & 10 & 20 & 90 & 364 & 1759 & 1953 & 11374 & 13665 \\
18000 & 10 & 20 & 90 & 366 & 1781 & 1988 & 12006 & 14446 \\
19000 & 10 & 20 & 90 & 369 & 1800 & 2002 & 12651 & 15236 \\
20000 & 10 & 20 & 90 & 369 & 1828 & 2032 & 13310 & 16033 \\
21000 & 10 & 20 & 90 & 369 & 1853 & 2054 & 13968 & 16824 \\
\end{tabular}
\caption{Number of Cuts per Iteration for the Various Bundling Schemes.}
\label{tab:cuts}
\end{table}

Figure \ref{fig:agg_100} presents the results comparing the cut
aggregation schemes on the small instances, while Figure
\ref{fig:errors} depicts the behavior of three bundling strategies,
one cut, hub aggregation, and leg aggregation, on the large
instances. They both present the average runtime per instance; In
addition, Figure \ref{fig:errors} also gives the standard deviation in
the form of an error bar (a dot on the graph indicates a standard
deviation of more than 100s). Figure \ref{fig:iterations} gives the
number of iterations for all bundling schemes on the large
instances. Table \ref{tab:cuts} also presents the number of cuts
generated per iteration for each bundling scheme.

Leg bundling is the best performing scheme: It is both more efficient
and more stable as the size of the instances grows. A probable
explanation is that the bus leg in the trip filtering clusters trips
with near-identical features with respect to their network usage (same
initial and final hubs). As a consequence, the resulting cuts generate
efficient partitions of the solution space. These results concur with
those presented by \citet{DeCamargo2008} who showed that the
computational overhead of generating a cut per commodity far outweighs
the potential gain in iterations and those by \citet{contreras2011benders} who
aggregate cuts by hubs with encouraging results.

\subsubsection{Core Point Update}

\begin{figure}[th!]
    \centering
    \small
\begingroup
  \makeatletter
  \providecommand\color[2][]{%
    \GenericError{(gnuplot) \space\space\space\@spaces}{%
      Package color not loaded in conjunction with
      terminal option `colourtext'%
    }{See the gnuplot documentation for explanation.%
    }{Either use 'blacktext' in gnuplot or load the package
      color.sty in LaTeX.}%
    \renewcommand\color[2][]{}%
  }%
  \providecommand\includegraphics[2][]{%
    \GenericError{(gnuplot) \space\space\space\@spaces}{%
      Package graphicx or graphics not loaded%
    }{See the gnuplot documentation for explanation.%
    }{The gnuplot epslatex terminal needs graphicx.sty or graphics.sty.}%
    \renewcommand\includegraphics[2][]{}%
  }%
  \providecommand\rotatebox[2]{#2}%
  \@ifundefined{ifGPcolor}{%
    \newif\ifGPcolor
    \GPcolortrue
  }{}%
  \@ifundefined{ifGPblacktext}{%
    \newif\ifGPblacktext
    \GPblacktextfalse
  }{}%
  \let\gplgaddtomacro\g@addto@macro
  \gdef\gplbacktext{}%
  \gdef\gplfronttext{}%
  \makeatother
  \ifGPblacktext
    \def\colorrgb#1{}%
    \def\colorgray#1{}%
  \else
    \ifGPcolor
      \def\colorrgb#1{\color[rgb]{#1}}%
      \def\colorgray#1{\color[gray]{#1}}%
      \expandafter\def\csname LTw\endcsname{\color{white}}%
      \expandafter\def\csname LTb\endcsname{\color{black}}%
      \expandafter\def\csname LTa\endcsname{\color{black}}%
      \expandafter\def\csname LT0\endcsname{\color[rgb]{1,0,0}}%
      \expandafter\def\csname LT1\endcsname{\color[rgb]{0,1,0}}%
      \expandafter\def\csname LT2\endcsname{\color[rgb]{0,0,1}}%
      \expandafter\def\csname LT3\endcsname{\color[rgb]{1,0,1}}%
      \expandafter\def\csname LT4\endcsname{\color[rgb]{0,1,1}}%
      \expandafter\def\csname LT5\endcsname{\color[rgb]{1,1,0}}%
      \expandafter\def\csname LT6\endcsname{\color[rgb]{0,0,0}}%
      \expandafter\def\csname LT7\endcsname{\color[rgb]{1,0.3,0}}%
      \expandafter\def\csname LT8\endcsname{\color[rgb]{0.5,0.5,0.5}}%
    \else
      \def\colorrgb#1{\color{black}}%
      \def\colorgray#1{\color[gray]{#1}}%
      \expandafter\def\csname LTw\endcsname{\color{white}}%
      \expandafter\def\csname LTb\endcsname{\color{black}}%
      \expandafter\def\csname LTa\endcsname{\color{black}}%
      \expandafter\def\csname LT0\endcsname{\color{black}}%
      \expandafter\def\csname LT1\endcsname{\color{black}}%
      \expandafter\def\csname LT2\endcsname{\color{black}}%
      \expandafter\def\csname LT3\endcsname{\color{black}}%
      \expandafter\def\csname LT4\endcsname{\color{black}}%
      \expandafter\def\csname LT5\endcsname{\color{black}}%
      \expandafter\def\csname LT6\endcsname{\color{black}}%
      \expandafter\def\csname LT7\endcsname{\color{black}}%
      \expandafter\def\csname LT8\endcsname{\color{black}}%
    \fi
  \fi
    \setlength{\unitlength}{0.0500bp}%
    \ifx\gptboxheight\undefined%
      \newlength{\gptboxheight}%
      \newlength{\gptboxwidth}%
      \newsavebox{\gptboxtext}%
    \fi%
    \setlength{\fboxrule}{0.5pt}%
    \setlength{\fboxsep}{1pt}%
\begin{picture}(7200.00,5040.00)%
    \gplgaddtomacro\gplbacktext{%
      \colorrgb{0.50,0.50,0.50}%
      \put(814,857){\makebox(0,0)[r]{\strut{}$0$}}%
      \colorrgb{0.50,0.50,0.50}%
      \put(814,1641){\makebox(0,0)[r]{\strut{}$50$}}%
      \colorrgb{0.50,0.50,0.50}%
      \put(814,2424){\makebox(0,0)[r]{\strut{}$100$}}%
      \colorrgb{0.50,0.50,0.50}%
      \put(814,3208){\makebox(0,0)[r]{\strut{}$150$}}%
      \colorrgb{0.50,0.50,0.50}%
      \put(814,3991){\makebox(0,0)[r]{\strut{}$200$}}%
      \colorrgb{0.50,0.50,0.50}%
      \put(814,4775){\makebox(0,0)[r]{\strut{}$250$}}%
      \colorrgb{0.50,0.50,0.50}%
      \put(946,725){\rotatebox{-45}{\makebox(0,0)[l]{\strut{}100}}}%
      \colorrgb{0.50,0.50,0.50}%
      \put(1234,725){\rotatebox{-45}{\makebox(0,0)[l]{\strut{}200}}}%
      \colorrgb{0.50,0.50,0.50}%
      \put(1521,725){\rotatebox{-45}{\makebox(0,0)[l]{\strut{}300}}}%
      \colorrgb{0.50,0.50,0.50}%
      \put(1809,725){\rotatebox{-45}{\makebox(0,0)[l]{\strut{}400}}}%
      \colorrgb{0.50,0.50,0.50}%
      \put(2096,725){\rotatebox{-45}{\makebox(0,0)[l]{\strut{}500}}}%
      \colorrgb{0.50,0.50,0.50}%
      \put(2384,725){\rotatebox{-45}{\makebox(0,0)[l]{\strut{}600}}}%
      \colorrgb{0.50,0.50,0.50}%
      \put(2671,725){\rotatebox{-45}{\makebox(0,0)[l]{\strut{}700}}}%
      \colorrgb{0.50,0.50,0.50}%
      \put(2959,725){\rotatebox{-45}{\makebox(0,0)[l]{\strut{}800}}}%
      \colorrgb{0.50,0.50,0.50}%
      \put(3246,725){\rotatebox{-45}{\makebox(0,0)[l]{\strut{}900}}}%
      \colorrgb{0.50,0.50,0.50}%
      \put(3534,725){\rotatebox{-45}{\makebox(0,0)[l]{\strut{}1000}}}%
      \colorrgb{0.50,0.50,0.50}%
      \put(3821,725){\rotatebox{-45}{\makebox(0,0)[l]{\strut{}1100}}}%
      \colorrgb{0.50,0.50,0.50}%
      \put(4109,725){\rotatebox{-45}{\makebox(0,0)[l]{\strut{}1200}}}%
      \colorrgb{0.50,0.50,0.50}%
      \put(4396,725){\rotatebox{-45}{\makebox(0,0)[l]{\strut{}1300}}}%
      \colorrgb{0.50,0.50,0.50}%
      \put(4684,725){\rotatebox{-45}{\makebox(0,0)[l]{\strut{}1400}}}%
      \colorrgb{0.50,0.50,0.50}%
      \put(4971,725){\rotatebox{-45}{\makebox(0,0)[l]{\strut{}1500}}}%
      \colorrgb{0.50,0.50,0.50}%
      \put(5259,725){\rotatebox{-45}{\makebox(0,0)[l]{\strut{}1600}}}%
      \colorrgb{0.50,0.50,0.50}%
      \put(5546,725){\rotatebox{-45}{\makebox(0,0)[l]{\strut{}1700}}}%
      \colorrgb{0.50,0.50,0.50}%
      \put(5834,725){\rotatebox{-45}{\makebox(0,0)[l]{\strut{}1800}}}%
      \colorrgb{0.50,0.50,0.50}%
      \put(6121,725){\rotatebox{-45}{\makebox(0,0)[l]{\strut{}1900}}}%
      \colorrgb{0.50,0.50,0.50}%
      \put(6409,725){\rotatebox{-45}{\makebox(0,0)[l]{\strut{}2000}}}%
      \colorrgb{0.50,0.50,0.50}%
      \put(6696,725){\rotatebox{-45}{\makebox(0,0)[l]{\strut{}2100}}}%
    }%
    \gplgaddtomacro\gplfronttext{%
      \csname LTb\endcsname%
      \put(176,2816){\rotatebox{-270}{\makebox(0,0){\strut{}Time in s}}}%
      \put(3850,154){\makebox(0,0){\strut{}Instance}}%
      \csname LTb\endcsname%
      \put(1933,4602){\makebox(0,0)[l]{\strut{}Split Pareto}}%
      \csname LTb\endcsname%
      \put(1933,4382){\makebox(0,0)[l]{\strut{}Core point update}}%
    }%
    \gplbacktext
    \put(0,0){\includegraphics{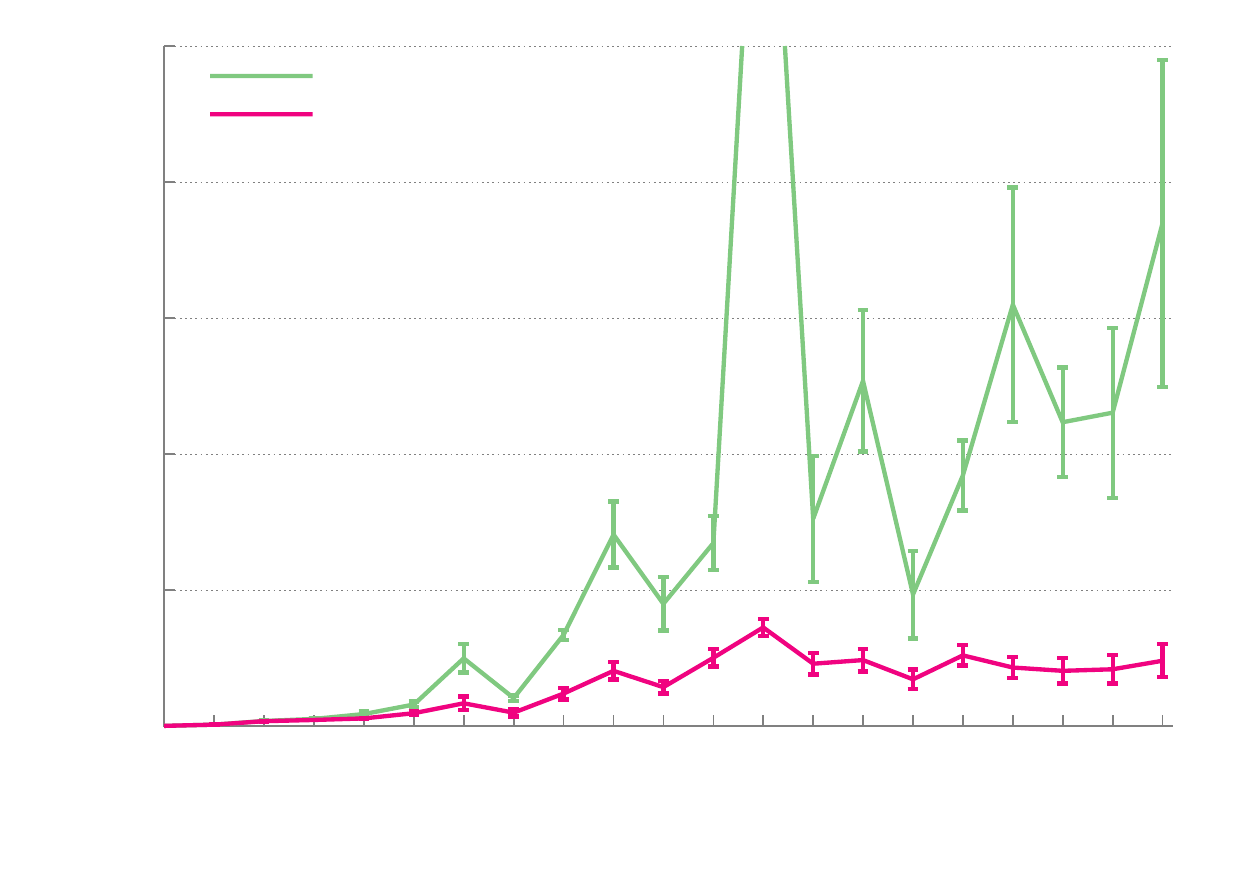}}%
    \gplfronttext
  \end{picture}%
\endgroup
    \caption{The Impact of the Core Point Updating Rule using $H_{20} \text{ \& } \alpha = 0.1$.}
    \label{fig:huberrors}
\end{figure}

Figure \ref{fig:huberrors} presents the benefits of the update rule
for the core point using the set of twenty potential hubs $H_{20}$ and
$\alpha = 0.1$. The key message is that the core point updates make
the Benders decomposition both faster and more stable.

\subsubsection{MIP versus Benders Decomposition}

\begin{figure}[th!]
    \centering
    \small
\begingroup
  \makeatletter
  \providecommand\color[2][]{%
    \GenericError{(gnuplot) \space\space\space\@spaces}{%
      Package color not loaded in conjunction with
      terminal option `colourtext'%
    }{See the gnuplot documentation for explanation.%
    }{Either use 'blacktext' in gnuplot or load the package
      color.sty in LaTeX.}%
    \renewcommand\color[2][]{}%
  }%
  \providecommand\includegraphics[2][]{%
    \GenericError{(gnuplot) \space\space\space\@spaces}{%
      Package graphicx or graphics not loaded%
    }{See the gnuplot documentation for explanation.%
    }{The gnuplot epslatex terminal needs graphicx.sty or graphics.sty.}%
    \renewcommand\includegraphics[2][]{}%
  }%
  \providecommand\rotatebox[2]{#2}%
  \@ifundefined{ifGPcolor}{%
    \newif\ifGPcolor
    \GPcolortrue
  }{}%
  \@ifundefined{ifGPblacktext}{%
    \newif\ifGPblacktext
    \GPblacktextfalse
  }{}%
  \let\gplgaddtomacro\g@addto@macro
  \gdef\gplbacktext{}%
  \gdef\gplfronttext{}%
  \makeatother
  \ifGPblacktext
    \def\colorrgb#1{}%
    \def\colorgray#1{}%
  \else
    \ifGPcolor
      \def\colorrgb#1{\color[rgb]{#1}}%
      \def\colorgray#1{\color[gray]{#1}}%
      \expandafter\def\csname LTw\endcsname{\color{white}}%
      \expandafter\def\csname LTb\endcsname{\color{black}}%
      \expandafter\def\csname LTa\endcsname{\color{black}}%
      \expandafter\def\csname LT0\endcsname{\color[rgb]{1,0,0}}%
      \expandafter\def\csname LT1\endcsname{\color[rgb]{0,1,0}}%
      \expandafter\def\csname LT2\endcsname{\color[rgb]{0,0,1}}%
      \expandafter\def\csname LT3\endcsname{\color[rgb]{1,0,1}}%
      \expandafter\def\csname LT4\endcsname{\color[rgb]{0,1,1}}%
      \expandafter\def\csname LT5\endcsname{\color[rgb]{1,1,0}}%
      \expandafter\def\csname LT6\endcsname{\color[rgb]{0,0,0}}%
      \expandafter\def\csname LT7\endcsname{\color[rgb]{1,0.3,0}}%
      \expandafter\def\csname LT8\endcsname{\color[rgb]{0.5,0.5,0.5}}%
    \else
      \def\colorrgb#1{\color{black}}%
      \def\colorgray#1{\color[gray]{#1}}%
      \expandafter\def\csname LTw\endcsname{\color{white}}%
      \expandafter\def\csname LTb\endcsname{\color{black}}%
      \expandafter\def\csname LTa\endcsname{\color{black}}%
      \expandafter\def\csname LT0\endcsname{\color{black}}%
      \expandafter\def\csname LT1\endcsname{\color{black}}%
      \expandafter\def\csname LT2\endcsname{\color{black}}%
      \expandafter\def\csname LT3\endcsname{\color{black}}%
      \expandafter\def\csname LT4\endcsname{\color{black}}%
      \expandafter\def\csname LT5\endcsname{\color{black}}%
      \expandafter\def\csname LT6\endcsname{\color{black}}%
      \expandafter\def\csname LT7\endcsname{\color{black}}%
      \expandafter\def\csname LT8\endcsname{\color{black}}%
    \fi
  \fi
    \setlength{\unitlength}{0.0500bp}%
    \ifx\gptboxheight\undefined%
      \newlength{\gptboxheight}%
      \newlength{\gptboxwidth}%
      \newsavebox{\gptboxtext}%
    \fi%
    \setlength{\fboxrule}{0.5pt}%
    \setlength{\fboxsep}{1pt}%
\begin{picture}(7200.00,5040.00)%
    \gplgaddtomacro\gplbacktext{%
      \colorrgb{0.50,0.50,0.50}%
      \put(588,2772){\makebox(0,0)[r]{\strut{}$0.01$}}%
      \colorrgb{0.50,0.50,0.50}%
      \put(588,3276){\makebox(0,0)[r]{\strut{}$0.1$}}%
      \colorrgb{0.50,0.50,0.50}%
      \put(588,3780){\makebox(0,0)[r]{\strut{}$1$}}%
      \colorrgb{0.50,0.50,0.50}%
      \put(588,4283){\makebox(0,0)[r]{\strut{}$10$}}%
      \colorrgb{0.50,0.50,0.50}%
      \put(588,4787){\makebox(0,0)[r]{\strut{}$100$}}%
    }%
    \gplgaddtomacro\gplfronttext{%
      \csname LTb\endcsname%
      \put(0,2679){\rotatebox{-270}{\makebox(0,0){\strut{}Time in s (log)}}}%
      \put(2087,2652){\makebox(0,0){\strut{}Small Instances}}%
      \csname LTb\endcsname%
      \put(1707,4614){\makebox(0,0)[l]{\strut{}MIP}}%
      \csname LTb\endcsname%
      \put(1707,4394){\makebox(0,0)[l]{\strut{}Benders}}%
    }%
    \gplgaddtomacro\gplbacktext{%
      \colorrgb{0.50,0.50,0.50}%
      \put(588,504){\makebox(0,0)[r]{\strut{}$0.1$}}%
      \colorrgb{0.50,0.50,0.50}%
      \put(588,1008){\makebox(0,0)[r]{\strut{}$1$}}%
      \colorrgb{0.50,0.50,0.50}%
      \put(588,1512){\makebox(0,0)[r]{\strut{}$10$}}%
      \colorrgb{0.50,0.50,0.50}%
      \put(588,2015){\makebox(0,0)[r]{\strut{}$100$}}%
      \colorrgb{0.50,0.50,0.50}%
      \put(588,2519){\makebox(0,0)[r]{\strut{}$1000$}}%
      \colorrgb{0.50,0.50,0.50}%
      \put(720,452){\rotatebox{-45}{\makebox(0,0)[l]{\strut{}100}}}%
      \colorrgb{0.50,0.50,0.50}%
      \colorrgb{0.50,0.50,0.50}%
      \put(994,452){\rotatebox{-45}{\makebox(0,0)[l]{\strut{}300}}}%
      \colorrgb{0.50,0.50,0.50}%
      \colorrgb{0.50,0.50,0.50}%
      \put(1267,452){\rotatebox{-45}{\makebox(0,0)[l]{\strut{}500}}}%
      \colorrgb{0.50,0.50,0.50}%
      \colorrgb{0.50,0.50,0.50}%
      \put(1541,452){\rotatebox{-45}{\makebox(0,0)[l]{\strut{}700}}}%
      \colorrgb{0.50,0.50,0.50}%
      \colorrgb{0.50,0.50,0.50}%
      \put(1814,452){\rotatebox{-45}{\makebox(0,0)[l]{\strut{}900}}}%
      \colorrgb{0.50,0.50,0.50}%
      \colorrgb{0.50,0.50,0.50}%
      \put(2088,452){\rotatebox{-45}{\makebox(0,0)[l]{\strut{}1100}}}%
      \colorrgb{0.50,0.50,0.50}%
      \colorrgb{0.50,0.50,0.50}%
      \put(2361,452){\rotatebox{-45}{\makebox(0,0)[l]{\strut{}1300}}}%
      \colorrgb{0.50,0.50,0.50}%
      \colorrgb{0.50,0.50,0.50}%
      \put(2635,452){\rotatebox{-45}{\makebox(0,0)[l]{\strut{}1500}}}%
      \colorrgb{0.50,0.50,0.50}%
      \colorrgb{0.50,0.50,0.50}%
      \put(2908,452){\rotatebox{-45}{\makebox(0,0)[l]{\strut{}1700}}}%
      \colorrgb{0.50,0.50,0.50}%
      \colorrgb{0.50,0.50,0.50}%
      \put(3182,452){\rotatebox{-45}{\makebox(0,0)[l]{\strut{}1900}}}%
      \colorrgb{0.50,0.50,0.50}%
      \colorrgb{0.50,0.50,0.50}%
      \put(3455,452){\rotatebox{-45}{\makebox(0,0)[l]{\strut{}2100}}}%
    }%
    \gplgaddtomacro\gplfronttext{%
      \csname LTb\endcsname%
      \put(2087,0){\makebox(0,0){\strut{}$H_{20}$}}%
    }%
    \gplgaddtomacro\gplbacktext{%
      \colorrgb{0.50,0.50,0.50}%
      \put(3972,2772){\makebox(0,0)[r]{\strut{}$1$}}%
      \colorrgb{0.50,0.50,0.50}%
      \put(3972,3780){\makebox(0,0)[r]{\strut{}$10$}}%
      \colorrgb{0.50,0.50,0.50}%
      \put(3972,4787){\makebox(0,0)[r]{\strut{}$100$}}%
    }%
    \gplgaddtomacro\gplfronttext{%
      \csname LTb\endcsname%
      \put(5471,2652){\makebox(0,0){\strut{}Large Instances}}%
    }%
    \gplgaddtomacro\gplbacktext{%
      \colorrgb{0.50,0.50,0.50}%
      \put(3972,504){\makebox(0,0)[r]{\strut{}$10$}}%
      \colorrgb{0.50,0.50,0.50}%
      \put(3972,1176){\makebox(0,0)[r]{\strut{}$100$}}%
      \colorrgb{0.50,0.50,0.50}%
      \put(3972,1847){\makebox(0,0)[r]{\strut{}$1000$}}%
      \colorrgb{0.50,0.50,0.50}%
      \put(3972,2519){\makebox(0,0)[r]{\strut{}$10000$}}%
      \colorrgb{0.50,0.50,0.50}%
      \put(4104,452){\rotatebox{-45}{\makebox(0,0)[l]{\strut{}1000}}}%
      \colorrgb{0.50,0.50,0.50}%
      \colorrgb{0.50,0.50,0.50}%
      \put(4378,452){\rotatebox{-45}{\makebox(0,0)[l]{\strut{}3000}}}%
      \colorrgb{0.50,0.50,0.50}%
      \colorrgb{0.50,0.50,0.50}%
      \put(4651,452){\rotatebox{-45}{\makebox(0,0)[l]{\strut{}5000}}}%
      \colorrgb{0.50,0.50,0.50}%
      \colorrgb{0.50,0.50,0.50}%
      \put(4925,452){\rotatebox{-45}{\makebox(0,0)[l]{\strut{}7000}}}%
      \colorrgb{0.50,0.50,0.50}%
      \colorrgb{0.50,0.50,0.50}%
      \put(5198,452){\rotatebox{-45}{\makebox(0,0)[l]{\strut{}9000}}}%
      \colorrgb{0.50,0.50,0.50}%
      \colorrgb{0.50,0.50,0.50}%
      \put(5472,452){\rotatebox{-45}{\makebox(0,0)[l]{\strut{}11000}}}%
      \colorrgb{0.50,0.50,0.50}%
      \colorrgb{0.50,0.50,0.50}%
      \put(5745,452){\rotatebox{-45}{\makebox(0,0)[l]{\strut{}13000}}}%
      \colorrgb{0.50,0.50,0.50}%
      \colorrgb{0.50,0.50,0.50}%
      \put(6019,452){\rotatebox{-45}{\makebox(0,0)[l]{\strut{}15000}}}%
      \colorrgb{0.50,0.50,0.50}%
      \colorrgb{0.50,0.50,0.50}%
      \put(6292,452){\rotatebox{-45}{\makebox(0,0)[l]{\strut{}17000}}}%
      \colorrgb{0.50,0.50,0.50}%
      \colorrgb{0.50,0.50,0.50}%
      \put(6566,452){\rotatebox{-45}{\makebox(0,0)[l]{\strut{}19000}}}%
      \colorrgb{0.50,0.50,0.50}%
      \colorrgb{0.50,0.50,0.50}%
      \put(6839,452){\rotatebox{-45}{\makebox(0,0)[l]{\strut{}21000}}}%
    }%
    \gplgaddtomacro\gplfronttext{%
      \csname LTb\endcsname%
      \put(5471,0){\makebox(0,0){\strut{}$H_{20}$}}%
    }%
    \gplbacktext
    \put(0,0){\includegraphics{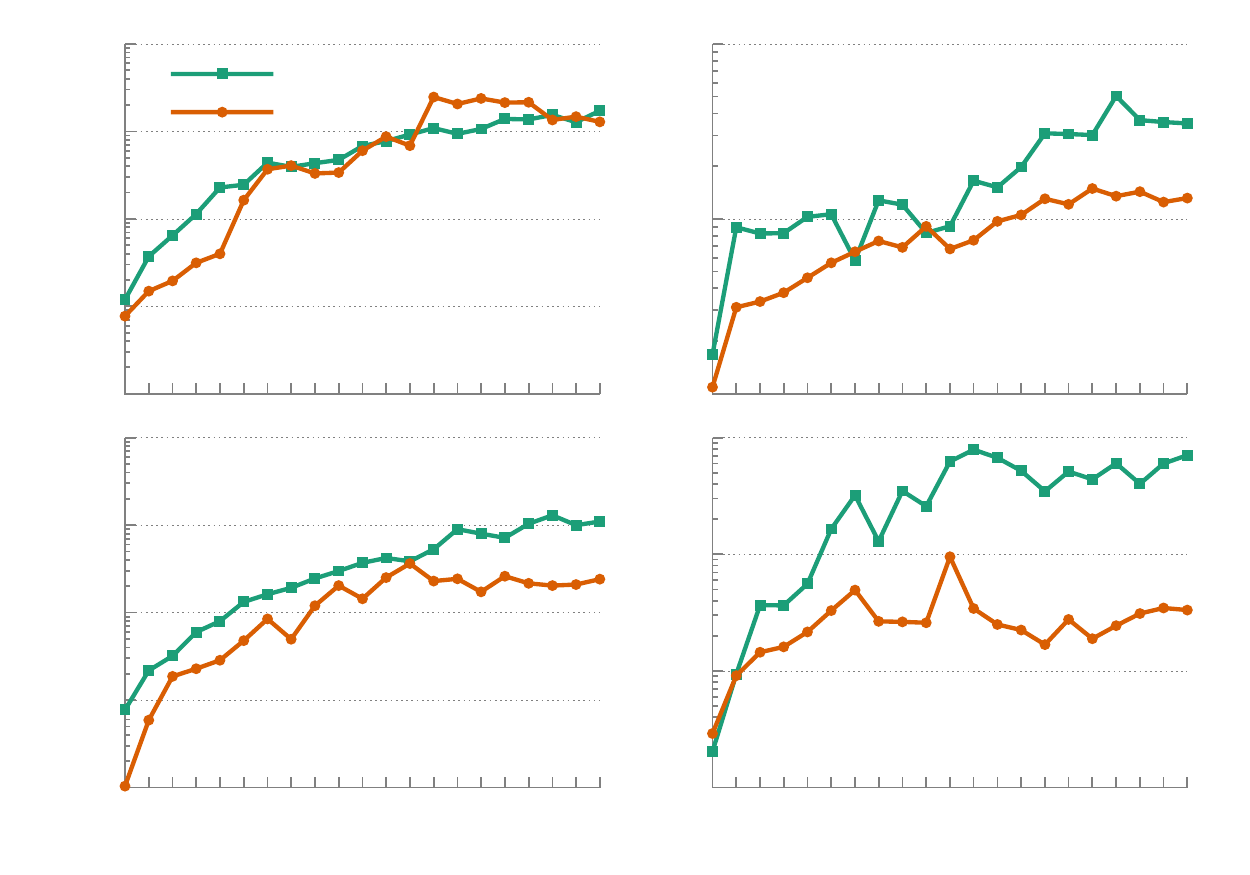}}%
    \gplfronttext
  \end{picture}%
\endgroup
    \caption{Comparing the MIP Model and the Final Benders Decomposition.}
    \label{fig:compare}
\end{figure}

Figure \ref{fig:compare} compares the standard MIP model against the
final version of our Benders decomposition algorithm (independent
Pareto sub-problems, leg aggregation, core point update). The
experiments use $\alpha = 0.1$ as it gives the most \emph{realistic}
results. The first column presents the results on the set of small
instances and the second column on the large instances. The first line
uses a configuration with ten hubs ($H_{10}$) and the second line
twenty hubs ($H_{20}$). As the instances grow larger, the gap between
the standard MIP and the Benders decomposition becomes increasingly
pronounced: Benders decomposition is almost two orders of magnitude
faster on the largest configuration.

\subsection{Benefits on the Case Study}

\begin{table}[t]%
\centering
\begin{tabular}{l *{5}{c}}%
\hline
\multirow{2}{*}{Day} & \multicolumn{3}{c}{{\sc BusPlus}} & \multicolumn{2}{c}{{\sc Action}} \\
\cmidrule(lr){2-4} \cmidrule(lr){5-6}
 & Buses (\$) & Cost (\$) & Time (s) & Cost (\$) & Time (s) \\
\midrule
Monday    & 45728.57 & 369420.37 & 829.88 & 402006.75 & 1635.22 \\
Tuesday   & 45728.57 & 362746.82 & 828.77 & 402006.75 & 1635.10 \\
Wednesday & 46436.58 & 372214.03 & 828.55 & 402006.75 & 1620.79 \\
Thursday  & 45899.13 & 376147.06 & 825.21 & 402006.75 & 1632.79 \\
Friday    & 43893.83 & 350709.85 & 819.64 & 402006.75 & 1610.85 \\
\midrule
\end{tabular}
\caption{Time and Cost Comparison Between BusPlus and Action.}%
\label{tab:network}%
\end{table}

We conclude this section by evaluating the approach on the real
case-study: The public transit system in Canberra.  Table
\ref{tab:network} compares the results of our {\sc BusPlus} approach
against the current public transportation system of Canberra known as
{\sc Action}. For {\sc BusPlus}, the table reports the cost of the bus
network, the total cost of the system, and the average travel time in
the three columns. For {\sc Action}, the table reports the cost of the
system and the average travel time. The experimental settings differ
slightly between {\sc BusPlus} and {\sc Action} to reflect the state
of the system as accurately as possible. In particular, we do not
include the waiting time for boarding buses in {\sc Action}, since
this data is not available.  We also set $\alpha$ to $0.1$ as it seems
to give the most realistic networks in terms of the balance between
costs and convenience.

The results show that the {\sc BusPlus} approach divides the average
travel time by two, even when taking into account the bus waiting
time. The main reason for this reduction is the significant number of
trips being served by taxis only. Moreover, even though the main
expense of {\sc BusPlus} comes from the taxi trips, the overall cost
of the system is about ten percent lower than the cost of {\sc
  Action}. This is an interesting result, since the taxi costs here
are intentionally exaggerated and correspond to existing full fares.

\begin{figure}[tb]
\centering
    \begin{subfigure}{0.45\columnwidth}
        \includemap{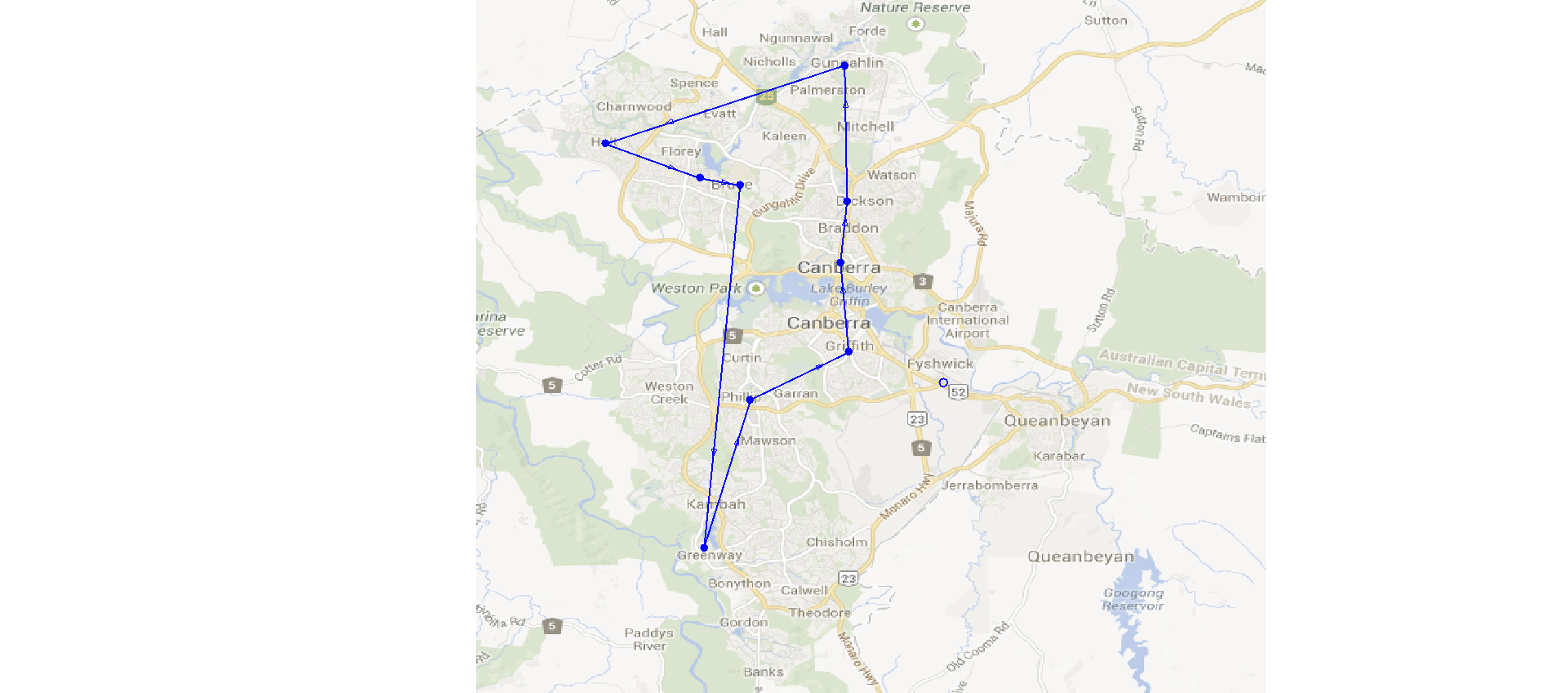}
        \subcaption{$H_{10}$ \& $\alpha = 10^{-5}$}
        \label{fig:h10_eps}
    \end{subfigure}%
    \hfill
    \begin{subfigure}{0.45\columnwidth}
        \includemap{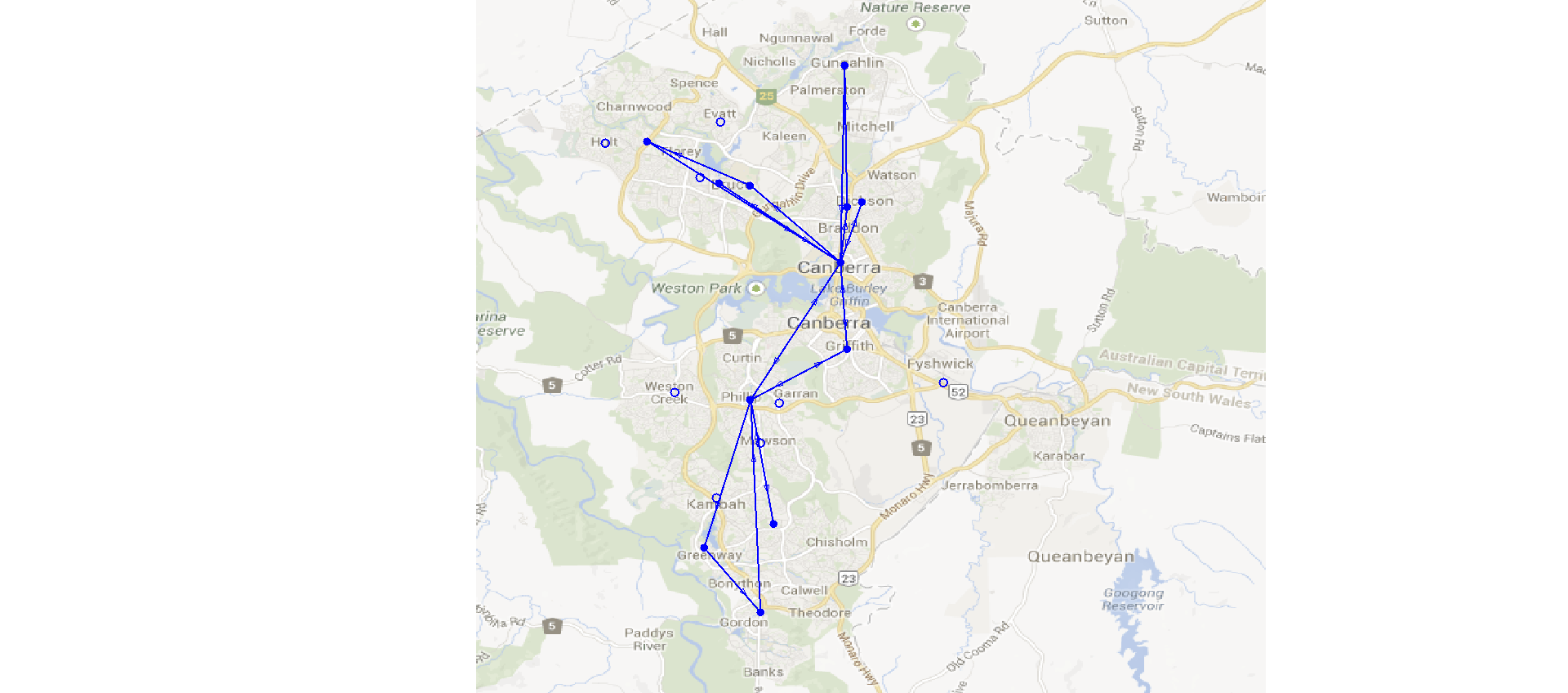}
        \subcaption{$H_{20}$ \& $\alpha = 0.1$}
        \label{fig:h20_01}
    \end{subfigure}
    \caption{Example networks using geographical data}
\label{fig:networks}
\end{figure}

Figure \ref{fig:networks} displays the networks obtained for the
$H_{10}$ and $H_{20}$ configurations. The $H_{20}$ configurations is
particularly interesting: It is organized around two interconnected
centers, which serve some of the most densely populated areas with
short. flower-like routes. 

\section{Conclusion}

The \busplus{} project aims at improving the off-peak hours public
transit service in Canberra, Australia. The city covers a wide
geographical area which makes public transportation particularly
challenging. To address the difficulty, the \busplus{} project
proposes a hub and shuttle model consisting of a combination of a few
high-frequency bus routes between key hubs and a large number of
shuttles that bring passengers from their origin to the closest hub
and take them from their last bus stop to their destination.

This paper focused on the design of bus network, which can be modeled
as a variant of the hub-arc location problem. It proposed a number of
pre-processing techniques, trip filtering and link filtering, to
reduce the problem size. Then it introduced a Benders decomposition
approach that uses dedicated solution techniques for solving
independent sub-problems, Pareto optimal cuts, cut bundling, and core
point update. The benefits of these design decisions are validated
using real-world data from public transit system in Canberra. The
Benders decomposition outperforms the natural MIP formulation by two
orders of magnitude. Moreover, the paper showed that the hub and
shuttle model may decrease transit time by a factor of 2, while
staying within the costs of the existing transit system.

There are several directions for future work, including the use of
parallel computation for the subproblems, the merging of the hub
selection and the network design in a single optimization problem, and
the ability to take into account of the scheduling, routing, and fleet
sizing aspects. 

It is also useful to mention that the \busplus{} project will undergo
live trial in the near future.

\end{document}